\documentclass{article}

     \usepackage[nonatbib, final]{neurips_2020}

\usepackage[utf8]{inputenc} %
\usepackage[T1]{fontenc}    %
\usepackage{hyperref}       %
\usepackage{url}            %
\usepackage{booktabs}       %
\usepackage{amsfonts}       %
\usepackage{nicefrac}       %
\usepackage{microtype}      %

\usepackage{xspace}

\makeatletter
\DeclareRobustCommand\onedot{\futurelet\@let@token\@onedot}
\def\@onedot{\ifx\@let@token.\else.\null\fi\xspace}

\def\eg{\emph{e.g}\onedot} 
\def\ie{\emph{i.e}\onedot} 
\def\cf{\emph{c.f}\onedot} 
 
\def\wrt{w.r.t\onedot} 

\makeatother

\usepackage{times}
\usepackage{epsfig}
\usepackage{graphicx}
\usepackage{amsmath}
\usepackage{amssymb}

\usepackage{xcolor} %

\usepackage{csquotes} %
\usepackage{amssymb} %
\usepackage{mathtools} %
\usepackage{placeins} %
\usepackage{bbold}
\usepackage{breqn}
\usepackage{graphbox} %
\usepackage{multirow}
\usepackage[font=small]{caption}
\usepackage[font=small]{subcaption} %
\usepackage{booktabs} %
\usepackage{url}
\usepackage{makecell}
\usepackage{nicefrac}

\title{Network-to-Network Translation with \\Conditional Invertible Neural
Networks}

\author{%
  Robin Rombach\thanks{Both authors contributed equally to this
  work.\newline
  Code available at \url{https://github.com/CompVis/net2net}.}
  \qquad\qquad Patrick Esser\footnotemark[1] \qquad\qquad Bj\"orn Ommer \\
  IWR, HCI, Heidelberg University \\
  \texttt{firstname.lastname@iwr.uni-heidelberg.de}
}

\newcommand{\oneenc}{\Phi}
\newcommand{\onedec}{\Psi}
\newcommand{\twoenc}{\Theta}
\newcommand{\twodec}{\Lambda}
\newcommand{\modelone}{f}
\newcommand{\modeltwo}{g}
\newcommand{\x}{x}
\newcommand{\y}{y}
\newcommand{\domainone}{\mathcal{D}_{\x}}
\newcommand{\domaintwo}{\mathcal{D}_{\y}}

\newcommand{\condinn}{\tau}
\newcommand{\modelrep}{z}
\newcommand{\repone}{\modelrep_{\oneenc}}
\newcommand{\reptwo}{\modelrep_{\twoenc}}
\newcommand{\normaldistr}{\mathcal{N}}
\newcommand{\id}{\mathbb{1}}

\newcommand{\modelinv}{v}
\newcommand{\aerep}{\bar{z}}

\newcommand{\xrec}{\bar{x}}

\newcommand{\compose}{\circ}

\newcommand{\loss}{\mathcal{L}}

\newcommand{\diag}{\operatorname{diag}}
\newcommand{\KL}{\operatorname{KL}}
\newcommand{\expect}{\mathbb{E}}

\newcommand{\RR}{\mathbb{R}}

\newcommand{\nimages}[1]{N^f}

\providecommand{\impath}[1]{}
\providecommand{\impatha}[1]{}
\providecommand{\impathb}[1]{}
\providecommand{\impathc}[1]{}
\providecommand{\impathd}[1]{}
\providecommand{\starpath}[1]{}
\providecommand{\imwidth}{}
\providecommand{\imwidtha}{}
\providecommand{\imwidthb}{}
\providecommand{\imheight}{}
\providecommand{\imwidthmult}[1]{}
\providecommand{\imwidthmulta}[1]{}
\providecommand{\imwidthmultb}[1]{}
\providecommand{\imwidthmultc}[1]{}

\newcommand{\creativityfigure}{
 \begin{figure*}[thbp]
   \renewcommand{\imwidth}{0.23\textwidth}
   \renewcommand{\imwidtha}{0.115\textwidth}
   \renewcommand{\impatha}[1]{img/portrait/##1}
   \renewcommand{\impathb}[1]{img/anime/##1}
   \renewcommand{\impathc}[1]{img/celebify/##1}
   \renewcommand{\impathd}[1]{img/mfhqanimalhumanswap/##1}
	\centering
	\begin{tabular}{c c c c c}
	\scriptsize{Oil-Portrait to Photography} &	\scriptsize{Anime to Photography} & \scriptsize{FFHQ to CelebA-HQ} & \multicolumn{2}{c}{\scriptsize{FFHQ to AFHQ}}\\
	\toprule
 	 	\vspace*{-0.2cm}
 	\includegraphics[width=\imwidth]{\impatha{newton}} \hspace*{-1.0em}& 
 	\includegraphics[width=\imwidth]{\impathb{15628_re}} \hspace*{-1.0em}& 
 	\includegraphics[width=\imwidth]{\impathc{10520}} \hspace*{-1.0em}&
    \includegraphics[width=\imwidtha]{\impathd{xdec_rec_000083}} \hspace*{-1.3em}&    
    \includegraphics[width=\imwidtha]{\impathd{xdec_swap_000083}} \\ 
 	 	\vspace*{-0.2cm}
 	\includegraphics[width=\imwidth]{\impatha{mona2real}} \hspace*{-1.0em}&
 	\includegraphics[width=\imwidth]{\impathb{15633_re}} \hspace*{-1.0em}& 
 	\includegraphics[width=\imwidth]{\impathc{11347}} \hspace*{-1.0em}&
    \includegraphics[width=\imwidtha]{\impathd{xdec_rec_000089}} \hspace*{-1.3em}&    
    \includegraphics[width=\imwidtha]{\impathd{xdec_swap_000089}} \\ 
	 	\vspace*{-0.1cm} 	
 	\includegraphics[width=\imwidth]{\impatha{gal2real}} \hspace*{-1.0em} & 	
 	\includegraphics[width=\imwidth]{\impathb{16368_re}} \hspace*{-1.0em} &
	\includegraphics[width=\imwidth]{\impathc{10618}}  \hspace*{-1.0em} &
    \includegraphics[width=\imwidtha]{\impathd{xdec_swap_000042}} \hspace*{-1.3em}&    
    \includegraphics[width=\imwidtha]{\impathd{xdec_rec_000042}} \\ 
	\end{tabular}
  \setlength{\belowcaptionskip}{-1.5em}
 	\caption{Unpaired Transfer: Bringing oil portraits and animes to live by
   projecting them onto the FFHQ dataset (column 1 and 2, respectively). Column
   3 visualizes the more subtle differences introduced when translating between
   different datasets of human faces such as FFHQ and CelebA-HQ.
   Column 4 shows a translation between the more diverse modalities of human and
   animal faces. See also Sec.~\ref{subsec:imgmod}~and~\ref{supp:unpaired}.}
 	\label{fig:secondpage}
 \end{figure*}
}

\newcommand{\superres}{
 \begin{figure*}[t!]
   \renewcommand{\imwidtha}{0.10\textwidth}
   \renewcommand{\imwidthb}{0.30\textwidth}
   \renewcommand{\impatha}[1]{img/superres/af16x16/##1}
   \renewcommand{\impathb}[1]{img/superres/faces32x32/##1}  
	\centering
	\begin{tabular}{c c   c c}
\scriptsize{$16 \times 16$}	& \scriptsize{to $256 \times 256$ Animalfaces} &
\scriptsize{$32 \times 32$}	& \scriptsize{to $256 \times 256$ CelebA-HQ/FFHQ} \\
	\toprule

 	\includegraphics[width=\imwidtha, align=c]{\impatha{88in}} & 
 		\hspace*{-1em}
 	\includegraphics[width=\imwidthb, align=c]{\impatha{88sample1}} &
 	
 	\includegraphics[width=\imwidtha, align=c]{\impathb{4243in}} & 
 	 	\hspace*{-1em}
 	\includegraphics[width=\imwidthb, align=c]{\impathb{4243sample3}} \\
 	
 	\includegraphics[width=\imwidtha, align=c]{\impatha{570in}} & 
 	 	\hspace*{-1em}
 	\includegraphics[width=\imwidthb, align=c]{\impatha{570sample2}} &
 	
 	\includegraphics[width=\imwidtha, align=c]{\impathb{11745in}} & 
 	 	\hspace*{-1em}
 	\includegraphics[width=\imwidthb, align=c]{\impathb{11745sample1}} \\
	
 	\includegraphics[width=\imwidtha, align=c]{\impatha{1491in}} & 
 	 	\hspace*{-1em}
 	\includegraphics[width=\imwidthb, align=c]{\impatha{1491sample1}} &
 	
 	\includegraphics[width=\imwidtha, align=c]{\impathb{13518in}} & 
 	 	\hspace*{-1em}
 	\includegraphics[width=\imwidthb, align=c]{\impathb{13518sample1}} \\

	\end{tabular}
  \setlength{\belowcaptionskip}{-1.5em}
 	\caption{Superresolution with Network-to-Network Translation. Here, we use our cINN to combine two autoencoders $\modelone$ and $\modeltwo$ to generatively combine two autoencoders living on image scales $32\times 32$ and $256 \times 256$.}
 	\label{fig:superres}
 \end{figure*}
}

\newcommand{\segtoimglogits}{
  \renewcommand{\impath}[1]{img/segtoimg/layerm2/##1}
  \renewcommand{\imwidth}{0.14\textwidth}
  \centering
  \begin{tabular}{c|cccc}
     $\oneenc(\x)$ & \multicolumn{4}{c}{\small{transferred} $\y = \twodec(\condinn(\modelinv \vert \oneenc(\x)))$} \\
    \hline
    \includegraphics[width=\imwidth, align=c]{\impath{segmentationmapsoft_000002}} &    
    \includegraphics[width=\imwidth, align=c]{\impath{sample0_000002}} & 
    \includegraphics[width=\imwidth, align=c]{\impath{sample5_000002}} & 
	\includegraphics[width=\imwidth, align=c]{\impath{sample3_000002}} & 
    \includegraphics[width=\imwidth, align=c]{\impath{sample4_000002}} \\

        \includegraphics[width=\imwidth, align=c]{\impath{segmentationmapsoft_000018}} &     
    \includegraphics[width=\imwidth, align=c]{\impath{sample0_000018}} & 
    \includegraphics[width=\imwidth, align=c]{\impath{sample5_000018}} & 
	\includegraphics[width=\imwidth, align=c]{\impath{sample3_000018}} & 
    \includegraphics[width=\imwidth, align=c]{\impath{sample4_000018}} \\

        \includegraphics[width=\imwidth, align=c]{\impath{segmentationmapsoft_000039}} &     
    \includegraphics[width=\imwidth, align=c]{\impath{sample0_000039}} & 
    \includegraphics[width=\imwidth, align=c]{\impath{sample5_000039}} & 
	\includegraphics[width=\imwidth, align=c]{\impath{sample3_000039}} & 
    \includegraphics[width=\imwidth, align=c]{\impath{sample4_000039}} \\

  \end{tabular}
  \caption{\emph{Logits} of segmentation expert.}
  \label{tab:segtoimglogits}
}

\newcommand{\segtoimgargmax}{
  \renewcommand{\impath}[1]{img/segtoimg/layerm1/##1}
  \renewcommand{\imwidth}{0.14\textwidth}
  \centering
  \begin{tabular}{c|cccc}
	$\oneenc(\x)$ & \multicolumn{4}{c}{\small{transferred} $\y = \twodec(\condinn(\modelinv \vert \oneenc(\x)))$} \\
    \hline
    \includegraphics[width=\imwidth, align=c]{\impath{segmentationmap_000002}} &    
    \includegraphics[width=\imwidth, align=c]{\impath{sample0_000002}} & 
    \includegraphics[width=\imwidth, align=c]{\impath{sample5_000002}} & 
	\includegraphics[width=\imwidth, align=c]{\impath{sample3_000002}} & 
    \includegraphics[width=\imwidth, align=c]{\impath{sample4_000002}} \\

        \includegraphics[width=\imwidth, align=c]{\impath{segmentationmap_000018}} &     
    \includegraphics[width=\imwidth, align=c]{\impath{sample0_000018}} & 
    \includegraphics[width=\imwidth, align=c]{\impath{sample5_000018}} & 
	\includegraphics[width=\imwidth, align=c]{\impath{sample3_000018}} & 
    \includegraphics[width=\imwidth, align=c]{\impath{sample4_000018}} \\

        \includegraphics[width=\imwidth, align=c]{\impath{segmentationmap_000039}} &     
    \includegraphics[width=\imwidth, align=c]{\impath{sample0_000039}} & 
    \includegraphics[width=\imwidth, align=c]{\impath{sample5_000039}} & 
	\includegraphics[width=\imwidth, align=c]{\impath{sample3_000039}} & 
    \includegraphics[width=\imwidth, align=c]{\impath{sample4_000039}} \\

  \end{tabular}
  \caption{\emph{Argmaxed logits} of segmentation expert.}
  \label{tab:segtoimgargmax}
}

\newcommand{\sobeltoimg}{
  \renewcommand{\impath}[1]{img/animale2i/##1}
  \renewcommand{\imwidth}{0.14\textwidth}
  \centering
  \begin{tabular}{c|cccc}
    input $\x$ & \multicolumn{4}{c}{\small{transferred} $\y = \twodec(\condinn(\modelinv \vert \oneenc(\x)))$} \\
    \hline

    \includegraphics[width=\imwidth, align=c]{\impath{animale2i_001_000}} & 
    \includegraphics[width=\imwidth, align=c]{\impath{animale2i_001_001}} & 
    \includegraphics[width=\imwidth, align=c]{\impath{animale2i_001_002}} & 
    \includegraphics[width=\imwidth, align=c]{\impath{animale2i_001_003}} & 
    \includegraphics[width=\imwidth, align=c]{\impath{animale2i_001_004}} \\

    \includegraphics[width=\imwidth, align=c]{\impath{animale2i_006_000}} & 
    \includegraphics[width=\imwidth, align=c]{\impath{animale2i_006_001}} & 
    \includegraphics[width=\imwidth, align=c]{\impath{animale2i_006_002}} & 
    \includegraphics[width=\imwidth, align=c]{\impath{animale2i_006_003}} & 
    \includegraphics[width=\imwidth, align=c]{\impath{animale2i_006_004}} \\

    \includegraphics[width=\imwidth, align=c]{\impath{animale2i_007_000}} & 
    \includegraphics[width=\imwidth, align=c]{\impath{animale2i_007_001}} & 
    \includegraphics[width=\imwidth, align=c]{\impath{animale2i_007_002}} & 
    \includegraphics[width=\imwidth, align=c]{\impath{animale2i_007_003}} & 
    \includegraphics[width=\imwidth, align=c]{\impath{animale2i_007_004}} \\

  \end{tabular}
  \caption{Edge-to-Image using stylized ResNet classifier.}
  \label{tab:sobeltoimg}
}

\newcommand{\inpainting}{
  \renewcommand{\impath}[1]{img/animalinpainting/##1}
  \renewcommand{\imwidth}{0.14\textwidth}
  \centering
  \begin{tabular}{c|cccc}
   input $\x$ & \multicolumn{4}{c}{\small{transferred} $\y = \twodec(\condinn(\modelinv \vert \oneenc(\x)))$} \\
    \hline

    \includegraphics[width=\imwidth, align=c]{\impath{animalinpainting_001_000}} & 
    \includegraphics[width=\imwidth, align=c]{\impath{animalinpainting_001_001}} & 
    \includegraphics[width=\imwidth, align=c]{\impath{animalinpainting_001_002}} & 
    \includegraphics[width=\imwidth, align=c]{\impath{animalinpainting_001_003}} & 
    \includegraphics[width=\imwidth, align=c]{\impath{animalinpainting_001_004}} \\

    \includegraphics[width=\imwidth, align=c]{\impath{animalinpainting_006_000}} & 
    \includegraphics[width=\imwidth, align=c]{\impath{animalinpainting_006_001}} & 
    \includegraphics[width=\imwidth, align=c]{\impath{animalinpainting_006_002}} & 
    \includegraphics[width=\imwidth, align=c]{\impath{animalinpainting_006_003}} & 
    \includegraphics[width=\imwidth, align=c]{\impath{animalinpainting_006_004}} \\

    \includegraphics[width=\imwidth, align=c]{\impath{animalinpainting_007_000}} & 
    \includegraphics[width=\imwidth, align=c]{\impath{animalinpainting_007_001}} & 
    \includegraphics[width=\imwidth, align=c]{\impath{animalinpainting_007_002}} & 
    \includegraphics[width=\imwidth, align=c]{\impath{animalinpainting_007_003}} & 
    \includegraphics[width=\imwidth, align=c]{\impath{animalinpainting_007_004}} \\

	\end{tabular}
  \caption{Inpainting using vanilla ResNet classifier.}
  \label{tab:inpainting}

}

\newcommand{\animaltable}{
  \begin{figure*}
    \setlength{\tabcolsep}{2pt}
    \centering
    \begin{subtable}[t]{0.49\linewidth}
      \segtoimgargmax
    \end{subtable}
    \begin{subtable}[t]{0.49\linewidth}
      \segtoimglogits
    \end{subtable}
    \begin{subtable}[b]{0.49\linewidth}
      \rule{0pt}{3ex}
      \sobeltoimg
    \end{subtable}
    \begin{subtable}[b]{0.49\linewidth}
      \inpainting
    \end{subtable}
    \caption{Different Image-to-Image translation tasks solved with a single
    AE $\modeltwo$ and different experts $\modelone$. \vspace{-0.35cm}}
    \label{fig:animaltable}
  \end{figure*}
}

\newcommand{\humananimalswapb}{
  \centering
  \renewcommand{\impath}[1]{img/animalswapcrop/##1}
  \renewcommand{\imheight}{3.8em}
  \setlength{\tabcolsep}{2pt}
  \begin{tabular}{cc}
    \footnotesize{$\oneenc(\x)$\;\;\;}\rotatebox{90}{\scriptsize{exemplar $\y$}}&
    \includegraphics[height=\imheight]{\impath{exemplar_000096}}%
    \includegraphics[height=\imheight]{\impath{exemplar_000097}}%
    \includegraphics[height=\imheight]{\impath{exemplar_000098}}%
    \includegraphics[height=\imheight]{\impath{exemplar_000099}} \\%

\toprule

    \includegraphics[height=\imheight, align=c]{\impath{ctxt_000099}} &     
    \includegraphics[height=\imheight, align=c]{\impath{x_ss_cross_000099}} \\

    \includegraphics[height=\imheight, align=c]{\impath{ctxt_000100}} &     
    \includegraphics[height=\imheight, align=c]{\impath{x_ss_cross_000100}} \\

    \includegraphics[height=\imheight, align=c]{\impath{ctxt_000101}} &     
    \includegraphics[height=\imheight, align=c]{\impath{x_ss_cross_000101}} \\

	\bottomrule
  \end{tabular}
   \caption{Exemplar-guided image-to-image translation.}
  \label{fig:animalswap}
}

\newcommand{\humananimalswap}{
  \begin{figure*}

    \label{tab:crossings}
    \begin{subtable}[t]{0.49\linewidth}
      \humananimalswapb
    \end{subtable}
      \hspace{1em}
    \begin{subtable}[t]{0.49\linewidth}
      \unsupswap
    \end{subtable}
    \caption{
      In (a),
      a segmentation representation
    $\oneenc(\x)$ is translated under
    the guidance of the residual $\modelinv=\condinn^{-1}(\twoenc(\y) \vert
    \oneenc(\y))$ obtained from exemplar $y$.
    In (b),
    $\oneenc$ is the same as $\twoenc$, but applied after a
    spatial deformation of its input such that $\condinn$ learns to extract
    a shape representation into $\modelinv$, which then controls the
    target shape.
    }
  \label{fig:swaps}
  \end{figure*}
}

\newcommand{\berttobigflat}{
\begin{figure}[htb]
  \setlength{\tabcolsep}{1pt}
  \renewcommand{\impath}[1]{img/berttobig/##1}
  \renewcommand{\imwidth}{0.11\textwidth}
  \centering
  \begin{tabular}{ c  c c c | c  c c c}
	source domain $\x$ & \multicolumn{3}{c|}{target domain $\y$} & 	source
    domain $\x$ & \multicolumn{3}{c}{target domain $\y$}  \\
    \toprule
    \midrule
    \tiny \shortstack{\emph{Two people on a paddle} \\ \emph{boat in the water}} & 
    \includegraphics[width=\imwidth, align=c]{\impath{example155_sample2}} &     
	\includegraphics[width=\imwidth, align=c]{\impath{example155_sample3}} &     
    \includegraphics[width=\imwidth, align=c]{\impath{example155_sample0}} &
    \tiny \shortstack{\emph{A close up of} \\ \emph{a plant with broccoli}} & 
    \includegraphics[width=\imwidth, align=c]{\impath{example170_sample0}} &     
	\includegraphics[width=\imwidth, align=c]{\impath{example170_sample3}} &     
    \includegraphics[width=\imwidth, align=c]{\impath{example170_sample4}} \\
    \midrule
    \tiny \shortstack{\emph{A close up of} \\ \emph{a clock on a wall}} & 
    \includegraphics[width=\imwidth, align=c]{\impath{example474_sample1}} &     
	\includegraphics[width=\imwidth, align=c]{\impath{example474_sample2}} &     
    \includegraphics[width=\imwidth, align=c]{\impath{example474_sample4}} &
     \tiny \shortstack{\emph{A fighter jet flying} \\ \emph{through a cloudy sky}} & 
    \includegraphics[width=\imwidth, align=c]{\impath{example260_sample0}} &     
	\includegraphics[width=\imwidth, align=c]{\impath{example260_sample1}} &     
    \includegraphics[width=\imwidth, align=c]{\impath{example260_sample2}} \\
    \midrule
        
    \tiny \shortstack{\emph{A black bear is} \\ \emph{walking through the woods}} & 
    \includegraphics[width=\imwidth, align=c]{\impath{example15_sample1}} &     
	\includegraphics[width=\imwidth, align=c]{\impath{example15_sample2}} &     
    \includegraphics[width=\imwidth, align=c]{\impath{example15_sample3}} &
    \tiny \shortstack{\emph{A glass of wine} \\ \emph{sitting on a table}} & 
    \includegraphics[width=\imwidth, align=c]{\impath{example292_sample1}} &     
	\includegraphics[width=\imwidth, align=c]{\impath{example292_sample2}} &     
    \includegraphics[width=\imwidth, align=c]{\impath{example292_sample0}} \\    
    \midrule
     \tiny \shortstack{\emph{A man riding skis} \\ \emph{down a snow covered slope}} & 
    \includegraphics[width=\imwidth, align=c]{\impath{example270_sample0}} &     
	\includegraphics[width=\imwidth, align=c]{\impath{example270_sample1}} &     
    \includegraphics[width=\imwidth, align=c]{\impath{example270_sample3}} &
    \tiny \shortstack{\emph{A wooden chair sitting} \\ \emph{in front of a chair}} & 
    \includegraphics[width=\imwidth, align=c]{\impath{example5_sample0}} &     
	\includegraphics[width=\imwidth, align=c]{\impath{example5_sample2}} &     
    \includegraphics[width=\imwidth, align=c]{\impath{example5_sample4}} \\
    \midrule    
    \tiny \shortstack{\emph{A group of horses pulling} \\ \emph{a carriage down a dirt}} & 
    \includegraphics[width=\imwidth, align=c]{\impath{example374_sample0}} &     
	\includegraphics[width=\imwidth, align=c]{\impath{example374_sample2}} &     
    \includegraphics[width=\imwidth, align=c]{\impath{example374_sample4}} &
    \tiny \shortstack{\emph{A car parked} \\ \emph{in front of a building}} & 
    \includegraphics[width=\imwidth, align=c]{\impath{example107_sample1}} &     
	\includegraphics[width=\imwidth, align=c]{\impath{example107_sample2}} &     
    \includegraphics[width=\imwidth, align=c]{\impath{example107_sample3}} \\    
    \midrule
    \bottomrule  
  \end{tabular}    
  \caption{\emph{BERT} \cite{devlin2018bert} to \emph{BigGAN}
  \cite{brock2018large} transfer: Additional examples, which demonstrate high diversity in synthesized outputs.}
  \label{fig:berttobigsupp}
\end{figure}
}

\newcommand{\berttobigfirstpage}{
\begin{figure}[htb]
  \setlength{\tabcolsep}{2pt}
  \renewcommand{\impath}[1]{img/berttobig/##1}
  \renewcommand{\imwidth}{0.08\textwidth}
  \centering
  \begin{tabular}{ r r r  c  c c c}

	\multicolumn{3}{c}{\multirow{3}{*}{\includegraphics[width=0.47\textwidth,
    align=c]{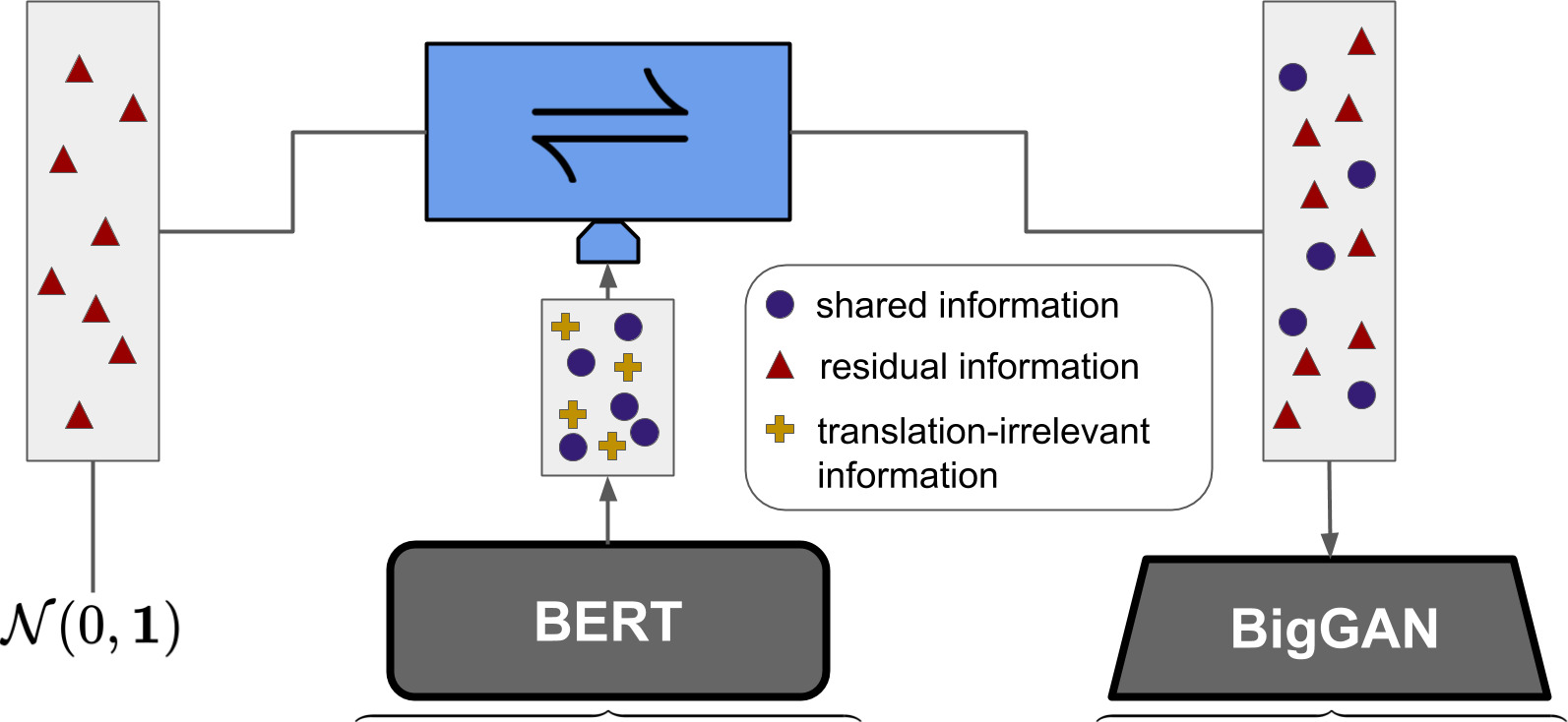}\hspace{2.2em}}} &
	source domain & 
    \multicolumn{3}{c}{target domain}  \\	

	\cmidrule{4-7}

    \multicolumn{3}{c}{\hspace{0.4em}} &
    \tiny \shortstack{\emph{A blue bird sitting} \\ \emph{on top of a field}} & 
    \includegraphics[width=\imwidth, align=c]{\impath{example10_sample1}} &     
	\includegraphics[width=\imwidth, align=c]{\impath{example10_sample2}} &     
    \includegraphics[width=\imwidth, align=c]{\impath{example10_sample3}} \\

		\cmidrule{4-7}    

    \multicolumn{3}{c}{\hspace{0.4em}} &
    \tiny \shortstack{\emph{A yellow bird is perched} \\ \emph{on a branch}} & 
    \includegraphics[width=\imwidth, align=c]{\impath{example116_sample1}} &     
	\includegraphics[width=\imwidth, align=c]{\impath{example116_sample2}} &     
    \includegraphics[width=\imwidth, align=c]{\impath{example116_sample3}} \\

    	\cmidrule{4-7}

      \hspace{4.2em}
      \tiny \shortstack{\emph{A park bench sitting} \\ \emph{in the middle of a
      park}} \hspace{0.3em} &
      &
      \includegraphics[width=\imwidth, align=c]{\impath{example315_sample1}}      
      \includegraphics[width=\imwidth, align=c]{\impath{example315_sample0}}
      \hspace{1.4525em} &     

      \tiny \shortstack{\emph{A couple of zebras} \\ \emph{are standing in a field}} & 
      \includegraphics[width=\imwidth, align=c]{\impath{example234_sample1}} &     
      \includegraphics[width=\imwidth, align=c]{\impath{example234_sample2}} &     
      \includegraphics[width=\imwidth, align=c]{\impath{example234_sample3}} \\

		\cmidrule{4-7}

      \hspace{4.2em}
    \tiny \shortstack{\emph{A pizza sitting on top} \\ \emph{of a white plate}}
    \hspace{0.3em} &
    &
    \includegraphics[width=\imwidth, align=c]{\impath{example263_sample1}}      
    \includegraphics[width=\imwidth, align=c]{\impath{example263_sample2}}
    \hspace{1.4525em} &     

    \tiny \shortstack{\emph{A school bus} \\ \emph{parked in a parking lot}} & 
    \includegraphics[width=\imwidth, align=c]{\impath{example128_sample1}} &     
    \includegraphics[width=\imwidth, align=c]{\impath{example128_sample2}} &     
    \includegraphics[width=\imwidth, align=c]{\impath{example128_sample0}} \\ 

  \end{tabular}    
  \caption{\emph{BERT} \cite{devlin2018bert} to \emph{BigGAN}
  \cite{brock2018large} transfer: Our approach enables translation between
  fixed off-the-shelve expert models such as BERT and BigGAN without having to
  modify or finetune them. \vspace*{-1em}}
  \label{fig:berttobig}
\end{figure}
}

\newcommand{\modelneurips}{
\begin{figure}[t]
\centering
\includegraphics[width=0.475\textwidth]{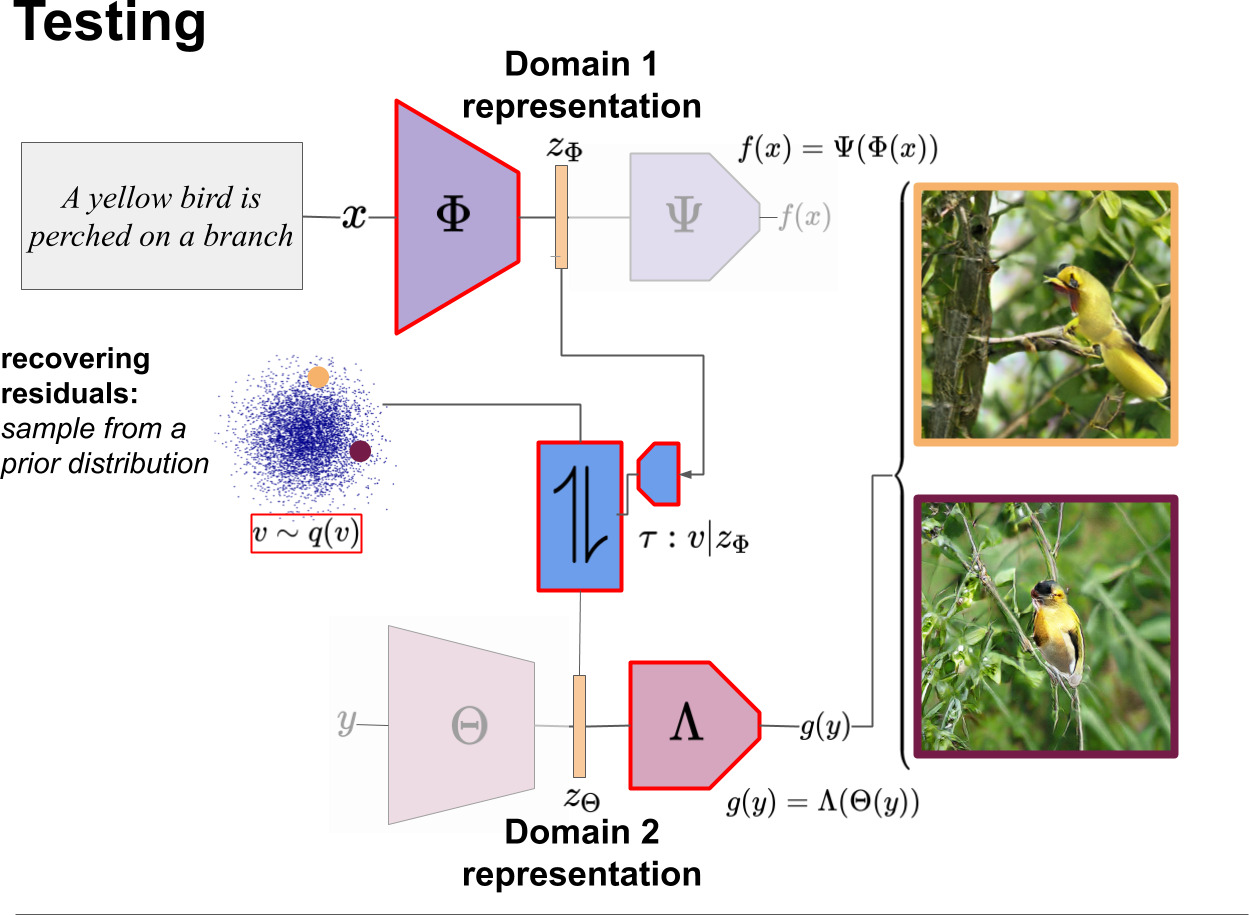}
\includegraphics[width=0.475\textwidth]{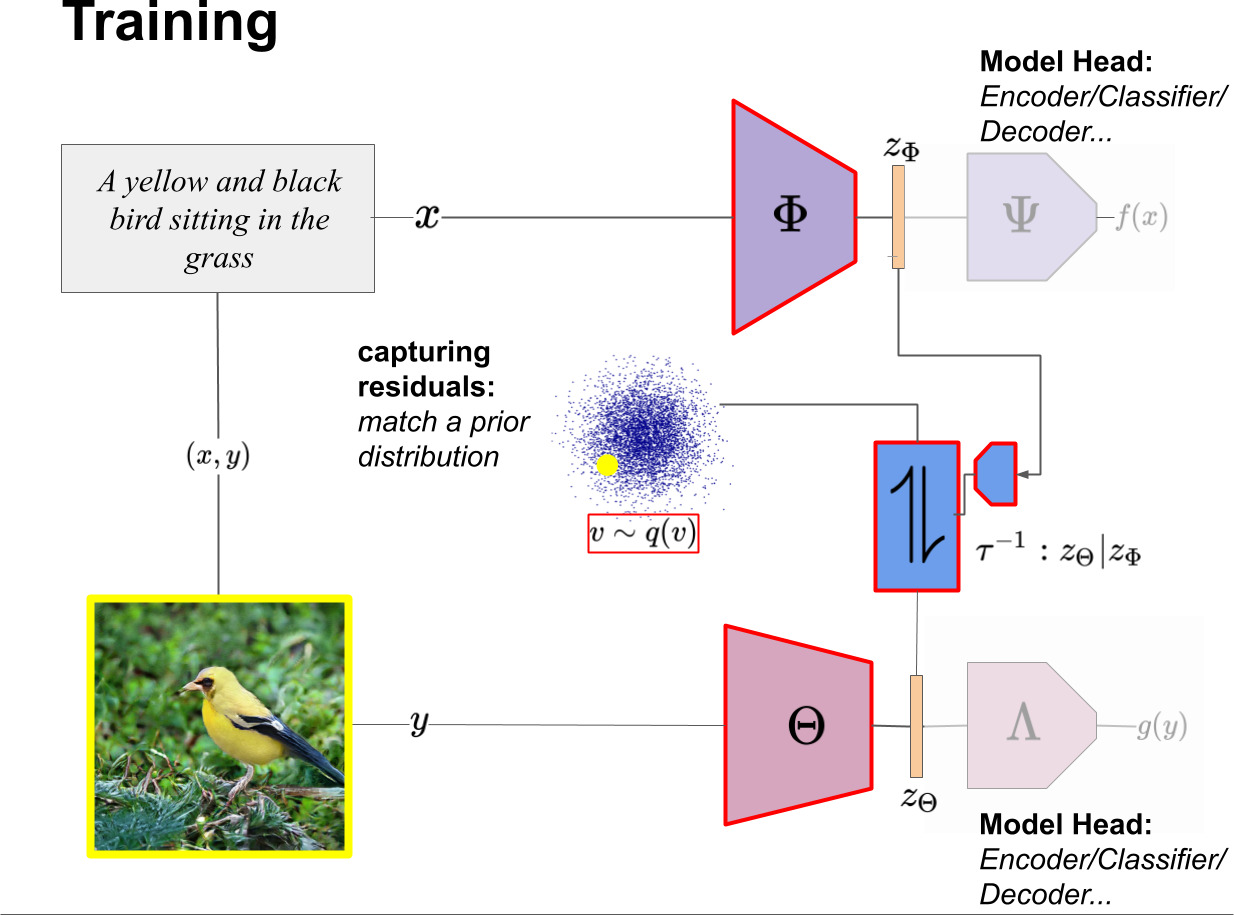}
\caption{Proposed architecture. We provide post-hoc model fusion for two given deep networks $\modelone = \oneenc \compose \onedec$ and $\modeltwo = \twoenc \compose \twodec$ which live on arbitrary domains $\domainone$ and $\domaintwo$. For deep representations $\repone = \oneenc(\x)$ and $\reptwo=\twoenc(\y)$, a conditional INN $\condinn$ learns to transfer between them by modelling the ambiguities \wrt the translation as an explicit residual, enabling transfer between given off-the-shelf models and their respective domains. \vspace{-0.3cm}}
\label{fig:modelnips} 
\end{figure}
}

\newcommand{\flowblock}{
\begin{figure}[htb]
\centering
\includegraphics[width=0.65\textwidth]{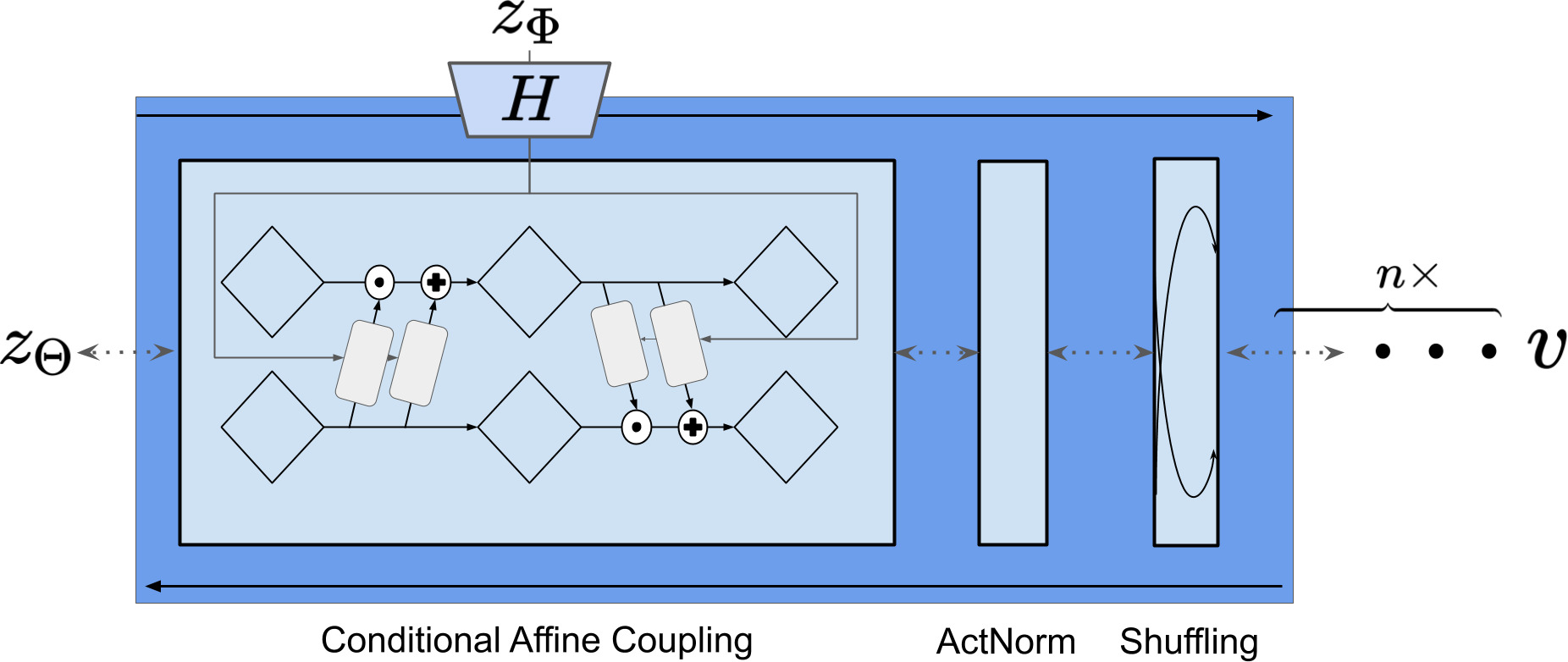}
\caption{A single (conditionally) invertible block used to build our cINN. We build the cINN from $n=20$ of these blocks for all of our experiments.}
  \label{fig:flowblock} \end{figure}
}

\newcommand{\celebtab}{  
\begin{figure*}  
  \setlength{\tabcolsep}{2pt}
  \renewcommand{\impath}[1]{img/celebaattr/##1}
  \renewcommand{\starpath}[1]{img/starganattr/##1}
  \renewcommand{\imwidth}{0.065\textwidth}
  \centering
  \begin{tabular}{ccccccccccc}
    \small{input} &
    \small{method} &
    \small{\emph{hair}} &
    \small{\emph{glasses}} &
    \small{\emph{gender}} &
    \phantom{abcdef} &
    \small{input} &
    \small{method} &
    \small{\emph{beard}} &
    \small{\emph{age}} &
    \small{\emph{smiling}}\\ 
    \cmidrule{1-5}\cmidrule{7-11}

    \multirow{2}{*}[-0.5em]{\includegraphics[width=\imwidth, align=c]{\impath{celebaattr_001_000}}} & 
    our &
    \includegraphics[width=\imwidth, align=c]{\impath{celebaattr_001_001}} & 
    \includegraphics[width=\imwidth, align=c]{\impath{celebaattr_001_002}} & 
    \includegraphics[width=\imwidth, align=c]{\impath{celebaattr_001_003}} & 
    &
    \multirow{2}{*}[-0.5em]{\includegraphics[width=\imwidth, align=c]{\impath{celebaattr_003_000}}} & 
    our &
    \includegraphics[width=\imwidth, align=c]{\impath{celebaattr_003_004}} & 
    \includegraphics[width=\imwidth, align=c]{\impath{celebaattr_003_005}} & 
    \includegraphics[width=\imwidth, align=c]{\impath{celebaattr_003_006}} \\
    & 
    \cite{choi2018stargan} &
    \includegraphics[width=\imwidth, align=c]{\starpath{stargan_001_001}} & 
    \includegraphics[width=\imwidth, align=c]{\starpath{stargan_001_002}} & 
    \includegraphics[width=\imwidth, align=c]{\starpath{stargan_001_003}} & 
    & & 
    \cite{choi2018stargan} &
    \includegraphics[width=\imwidth, align=c]{\starpath{stargan_003_004}} & 
    \includegraphics[width=\imwidth, align=c]{\starpath{stargan_003_005}} & 
    \includegraphics[width=\imwidth, align=c]{\starpath{stargan_003_006}} \\
    \cmidrule{1-5}\cmidrule{7-11}

    \multirow{2}{*}[-0.5em]{\includegraphics[width=\imwidth, align=c]{\impath{celebaattr_002_000}}} & 
    our &
    \includegraphics[width=\imwidth, align=c]{\impath{celebaattr_002_001}} & 
    \includegraphics[width=\imwidth, align=c]{\impath{celebaattr_002_002}} & 
    \includegraphics[width=\imwidth, align=c]{\impath{celebaattr_002_003}} & 
    &
    \multirow{2}{*}[-0.5em]{\includegraphics[width=\imwidth, align=c]{\impath{celebaattr_004_000}}} & 
    our &
    \includegraphics[width=\imwidth, align=c]{\impath{celebaattr_004_004}} & 
    \includegraphics[width=\imwidth, align=c]{\impath{celebaattr_004_005}} & 
    \includegraphics[width=\imwidth, align=c]{\impath{celebaattr_004_006}} \\
    & 
    \cite{choi2018stargan} &
    \includegraphics[width=\imwidth, align=c]{\starpath{stargan_002_001}} & 
    \includegraphics[width=\imwidth, align=c]{\starpath{stargan_002_002}} & 
    \includegraphics[width=\imwidth, align=c]{\starpath{stargan_002_003}} & 
    & & 
    \cite{choi2018stargan} &
    \includegraphics[width=\imwidth, align=c]{\starpath{stargan_004_004}} & 
    \includegraphics[width=\imwidth, align=c]{\starpath{stargan_004_005}} & 
    \includegraphics[width=\imwidth, align=c]{\starpath{stargan_004_006}} \\

    \cmidrule{1-5}\cmidrule{7-11}
    \cmidrule{1-5}\cmidrule{7-11}

    \multirow{2}{*}{FID} &
    our &
    \textbf{15.18} &
    \textbf{37.32} &
    \textbf{16.38} &
      &
    \multirow{2}{*}{FID} &
    our &
    \textbf{12.02} &
    \textbf{10.77} &
    \textbf{9.57} \\
    &
    \cite{choi2018stargan} &
    20.94 & 
    41.27 & 
    20.04 &
      &
    &
    \cite{choi2018stargan} &
    19.88 &
    21.77 &
    14.47 \\
    \cmidrule{1-5}\cmidrule{7-11}
  \end{tabular}
  \caption{
    We directly consider attribute vectors for $\repone$ to perform attribute
    modifications. We show both qualitative comparisons to
    \cite{choi2018stargan}, obtained by changing a single attribute of the
    input, as well as quantiative comparisons of FID scores, obtained after
    flipping a single attribute for all images of the test set. The results
    demonstrate the value of reusing a powerful, generic autoencoder (AE)
    $\modeltwo$ and repurposing it via our approach for a specific task, such
    as attribute modification, instead of learning an AE and the
    modification task simultaneously.
  }
  \label{fig:celebtab}
\end{figure*}  
}

\newcommand{\berttobiginception}{
  \begin{table}[b]
  \centering
  \caption{
    Inception and FID scores for BERT-to-BigGAN transfer on captions from
    COCO-stuff. Our approach is on-par with the current state of the
    art but does not require training of a text-encoder and image-decoder.}
      \begin{scriptsize}
     
  \begin{tabular}{cccccccc}
  \toprule
	 &
    \textbf{our} &
    SD-GAN \cite{yin2019semantics} &
	AttnGAN \cite{xu2018attngan} &
    StackGAN \cite{zhang2017stackgan} &
    DM-GAN  \cite{zhu2019dm}&
    MirrorGAN \cite{qiao2019mirrorgan}& 
	HDGAN  \cite{zhang2018photographic} \\
    \midrule
    IS $\uparrow$ &
    $\mathbf{34.7 \pm 0.3}$ &
    $\mathbf{35.7 \pm 0.5}$ &
    $25.9 \pm 0.5$ &
    $8.5 \pm 0.1$ &
    $30.5 \pm 0.6$ &
	$26.5 \pm 0.4$ & 
	$11.9 \pm 0.2$ \\
    FID $\downarrow$ &
    $\mathbf{30.63}$ &
    - &
    $35.49$ &
    - &
    $32.64$ &
	  - &
    - \\
	\bottomrule

  \end{tabular}
      \end{scriptsize}
  \label{tab:bertinception}
 \end{table}
}

\newcommand{\landscapeslayers}{
\begin{figure}[b!]
  \renewcommand{\impath}[1]{img/landscapelayers/##1}
  \renewcommand{\imwidth}{0.06\textwidth}
  \renewcommand{\imwidthmulta}{0.24\textwidth}
  \renewcommand{\imwidthmultb}{0.3\textwidth}
  \renewcommand{\imwidthmultc}{0.36\textwidth}
  \setlength{\tabcolsep}{2pt}
  \centering
  \begin{tabular}{ c  c  c  c}
    \footnotesize{$\x$} & \footnotesize{early layer: $\twodec(\reptwo)$} &
    \footnotesize{middle layer: $\twodec(\reptwo)$} & \footnotesize{last
    layer of $\modelone$: $\twodec(\reptwo)$} \\
    \toprule
	\includegraphics[width=\imwidth, align=c]{\impath{lm1/exemplar_000000}} &         
    \includegraphics[width=\imwidthmulta, align=c]{\impath{lm5/x_ss_sample_000000}} &     
    \includegraphics[width=\imwidthmultb, align=c]{\impath{lm2/x_ss_sample_000000}} &     
    \includegraphics[width=\imwidthmultc, align=c]{\impath{lm1/x_ss_sample_000000}} \\
    \midrule
    \includegraphics[width=\imwidth, align=c]{\impath{lm1/exemplar_000008}} &         
    \includegraphics[width=\imwidthmulta, align=c]{\impath{lm5/x_ss_sample_000008}} &     
    \includegraphics[width=\imwidthmultb, align=c]{\impath{lm2/x_ss_sample_000008}} &     
    \includegraphics[width=\imwidthmultc, align=c]{\impath{lm1/x_ss_sample_000008}} \\
    \midrule
    \includegraphics[width=\imwidth, align=c]{\impath{lm1/exemplar_000068}} &         
    \includegraphics[width=\imwidthmulta, align=c]{\impath{lm5/x_ss_sample_000068}} &     
    \includegraphics[width=\imwidthmultb, align=c]{\impath{lm2/x_ss_sample_000068}} &     
    \includegraphics[width=\imwidthmultc, align=c]{\impath{lm1/x_ss_sample_000068}} \\
    \midrule
    \includegraphics[width=\imwidth, align=c]{\impath{lm1/exemplar_000097}} &         
    \includegraphics[width=\imwidthmulta, align=c]{\impath{lm5/x_ss_sample_000097}} &     
    \includegraphics[width=\imwidthmultb, align=c]{\impath{lm2/x_ss_sample_000097}} &     
    \includegraphics[width=\imwidthmultc, align=c]{\impath{lm1/x_ss_sample_000097}} \\
    \bottomrule  
  \end{tabular}    
  \caption{Translating different layers of an expert model $\modelone$ to the
  representation of an autoencoder $\modeltwo$ reveals the learned invariances of
  $\modelone$ and thus provides diagnostic insights. Here,
  $\modelone$ is a segmentation model, while $\modeltwo$ is the same
  AE as in Sec.~\ref{subsec:reuse}. For $\repone =
  \oneenc(\x)$, obtained from different layers of $\modelone$, we sample $\reptwo$ as in
  Eq.~\eqref{eq:sample} and synthesize corresponding images $\twodec(\reptwo)$. \vspace{-0.6cm}}
  \label{fig:landscapeslayers}
\end{figure}
}

\newcommand{\aearchitecture}{
\begin{table}[htb]
          \centering
          \begin{subtable}[t]{.45\textwidth}
              \centering
              {\begin{tabular}[t]{c}
                  \toprule
                  \midrule
                  RGB image $x\in \mathbb{R}^{128 \times 128 \times 3}$ \\
                  \midrule
                  Conv down $\to \mathbb{R}^{64\times 64 \times 64}$\\
                  \midrule
                  Norm, ReLU, MaxPool $\to \mathbb{R}^{32 \times 32 \times 64}$ \\
                  \midrule
                  $3\times$ BottleNeck $\to \mathbb{R}^{32 \times 32 \times 256}$ \\
                  \midrule
                  $4\times$ BottleNeck down $\to \mathbb{R}^{16 \times 16 \times 512}$ \\
                  \midrule
                  $23\times$ BottleNeck down $\to \mathbb{R}^{8 \times 8 \times 1024}$ \\
                  \midrule
                  $3\times$ BottleNeck down $\to \mathbb{R}^{4 \times 4 \times 2048}$ \\
                  \midrule
                  AvgPool, FC $\mapsto (\mu, \sigma^2) \in
                  \mathbb{R}^{128}\times \mathbb{R}^{128}$ \\
                  \midrule
                  \bottomrule
              \end{tabular}}
              \caption{\label{tab:aeencoder} Encoder based on
              \textsl{Resnet-101}.}
          \end{subtable}
          \begin{subtable}[t]{.45\textwidth}
              \centering
              {\begin{tabular}[t]{c}
                  \toprule
                  \midrule
                  $\aerep\in \mathbb{R}^{128} \sim \mathcal{N}(\mu, \diag(\sigma^2))$ \\
                  $3\times$ (FC, LReLU) $\to \mathbb{R}^{256}$ \\
                  FC, Softmax $\to \mathbb{R}^{1000}$ \\
                  Embed $\mapsto h \in \mathbb{R}^{128}$ \\
                  \midrule
                  FC($\aerep$) $\to \mathbb{R}^{4\times 4\times 16\cdot96}$ \\
                  \midrule
                  ResBlock($\aerep$,$h$) up $\to \mathbb{R}^{8\times 8\times 16\cdot96}$ \\
                  \midrule
                  ResBlock($\aerep$,$h$) up $\to \mathbb{R}^{16\times 16\times 8\cdot96}$ \\
                  \midrule
                  ResBlock($\aerep$,$h$) up $\to \mathbb{R}^{32\times 32\times 4\cdot96}$ \\
                  \midrule
                  ResBlock($\aerep$,$h$) up $\to \mathbb{R}^{64\times 64\times 2\cdot96}$ \\
                  \midrule
                  Non-Local Block $\to \mathbb{R}^{64\times 64\times 2\cdot96}$\\
                  \midrule
                  ResBlock($\aerep$,$h$) up $\to \mathbb{R}^{64\times 64\times 96}$ \\
                  \midrule
                  Norm, ReLU, Conv up $\to \mathbb{R}^{128\times 128\times 3}$ \\
                  \midrule
                  Tanh $\mapsto \xrec\in\mathbb{R}^{128\times 128\times 3}$ \\
                  \midrule
                  \bottomrule
              \end{tabular}}
              \caption{\label{tab:aedecoder} Decoder based on \textsl{BigGAN}.}
          \end{subtable}
   \caption{\label{tab:aearchitecture} Autoencoder architecture for the
         \textsl{CelebA} and \textsl{Animals} datasets at
         resolution $128 \times 128$.}                
\end{table}
}

\newcommand{\energyusage}{
\begin{table}[htb]
\centering
\begin{tabular}{cccccc}
\toprule
\midrule
Model & Time [days] & Hardware & Energy [kWh] & Cost [EUR] & $CO_2$ [kg] \\
\midrule
\textbf{our cINN} & $\mathbf{\leq1}$ & 1 NVIDIA Titan X & \textbf{14.4} & \textbf{3.11} & \textbf{4.26} \\
\midrule
BigGAN \cite{brock2018large} & 15 & 8 NVIDIA V100 & 1260.0 & 272.16 & 372.96 \\
\midrule
FUNIT \cite{liu2019few} & 14 & 8 NVIDIA V100 & 1176.0 & 254.02 & 348.10 \\
\midrule
BERT \cite{devlin2018bert} & 10.3 & 8 NVIDIA V100 & 865.2 & 186.88 & 256.10 \\
\midrule
\bottomrule
\end{tabular}
\caption{
  Comparison of computational costs for a single training run of different
  models. Energy consumption of a Titan
  X is based on the recommended system power (0.6 kW) by
  NVIDIA\protect\footnotemark,
  and energy consumption of eight V100 on the power (3.5 kW) of a NVIDIA DGX-1
  system\protect\footnotemark.
  Costs are based on the average price of 0.216 EUR per kWh in the EU\protect\footnotemark,
  and CO2 emissions on the average emissions of 0.296 kg CO2 per kWh in the EU\protect\footnotemark.
}
\label{tab:energy}
\end{table}
}

\newcommand{\minimodelstab}{
\begin{table}[htbp]
\centering
\begin{subtable}[t]{0.485\textwidth}
\centering
 \begin{tabular}[t]{c}
	\toprule
	\midrule
	input $z \in \mathbb{R}^{d}$ \\
    \midrule
    (FC, LReLU) $\rightarrow \mathbb{R}^{8\cdot d}$\\
	\midrule
	$2\times$ (FC, LReLU) $\rightarrow \mathbb{R}^{8 \cdot d}$\\
	\midrule
    (FC, LReLU) $\rightarrow \mathbb{R}^{d}$\\
    \midrule
    \bottomrule
   \end{tabular}  
   \caption{\label{tab:subbasicfully} Basic fully connected architecture.}
\end{subtable}
\begin{subtable}[t]{0.485\textwidth}
\centering
  \begin{tabular}[t]{c}
	\toprule
	\midrule
	input $y \in \mathbb{R}^{c \times h \times w}$ \\
    \midrule
    $n \times$ Conv down, ActNorm, LReLU  \\
    $\rightarrow \RR^{64 \cdot n \times h/2^n \times w/2^n}$ \\
    \midrule      
    Flatten, FC $\rightarrow \RR^{d}$ \\
    \midrule
    \bottomrule
   \end{tabular}  
\caption{ \label{tab:subembeddertab} Embedding module $H$, see Eq.~\eqref{eq:condembed}.}
\end{subtable}
\caption{\label{tab:basicfullyandembedder} \emph{(a)}: Architecture of the
  subnetworks $s_{\theta}$ and $t_{\theta}$ used to build the normalizing flow
  described in Sec.~\ref{sec:approach},~\ref{subsec:cinnarch}. Leaky ReLU
  (LReLU) uses a slope parameter $\alpha = 0.01$. For the cINN from
  Sec.~\ref{subsec:transtobig}, $d=268$, while for Sec.~\ref{subsec:reuse} and
  Sec.~\ref{subsec:imgmod}, $d=128$.~\emph{(b)} Architecture of the embedding
  module $H$, which is used to pre-process arbitrarily sized conditioning
  information $y$ via $h=H(y)$. Here, $n$ denotes the number of downsampling
  steps. If the conditioning $y$ does not have spatial dimensionality, we
  replace the whole network by a simple feedforward-architecture as in
  Tab.~\ref{tab:subbasicfully}.}
\end{table}
}

\newcommand{\suppminivae}{
\begin{table}[htb]
 \centering
  \begin{tabular}[t]{c}
	\toprule
	\midrule
	Embedding $h\in \mathbb{R}^{128}$ \\
	\midrule
	$3\times$ (FC, LReLU) $\rightarrow \mathbb{R}^{4096}$\\
	\midrule
	(FC, LReLU) $\rightarrow \mathbb{R}^{128}$\\
	\midrule
	$\mu, \sigma^2$: for each: \\
	\midrule
	$3\times$ (FC, LReLU) $\rightarrow \mathbb{R}^{4096}$\\
	\midrule
	(FC, LReLU) $\rightarrow \mathbb{R}^{128}$\\
	\midrule
	$h \in \mathbb{R}^{128} \sim \normaldistr(\mu, \diag(\sigma^2))$ \\
    \midrule
    $4 \times $ (FC, LReLU) $\rightarrow \mathbb{R}^{4096}$\\
    \midrule
    (FC, LReLU) $\rightarrow \mathbb{R}^{128}$\\
	\midrule
    \bottomrule
   \end{tabular}  
\caption{\label{tab:minivae} 
Training a cINN on synthetic BigGAN data requires to dequantize the discrete class information which is used as conditioning information for the decoder. To this end, we make use of a variational autoencoder as described in Sec.~\ref{subsec:berttobigdetails}, which provides a stochastic reconstruction of its input $h$. For Leaky ReLU, we use a slope parameter of $\alpha=0.01$.}
\end{table}
}

\newcommand{\supphumananimalswap}{
\begin{figure}
  \centering
  \renewcommand{\impath}[1]{img/suppmfhqanimalhumanswap/##1}
  \renewcommand{\imheight}{3.8em}
  \setlength{\tabcolsep}{2pt}
  \begin{tabular}{cc}
    \toprule
    in & \includegraphics[height=\imheight, align=c]{\impath{swap-02}} \\
    out & \includegraphics[height=\imheight, align=c]{\impath{swap-06}} \\
    \midrule
    in & \includegraphics[height=\imheight, align=c]{\impath{swap-03}} \\
    out & \includegraphics[height=\imheight, align=c]{\impath{swap-07}} \\
    \bottomrule
  \end{tabular}
   \caption{Additional examples for unpaired translation between human and
   animal faces as in Fig.~\ref{fig:secondpage}. Our approach naturally
   provides translations in both directions (see Sec.~\ref{supp:unpaired}).
   Inputs are randomly choosen test examples from either the human or the
   animal data and translated to the respective other one.}
  \label{fig:suppswap}
\end{figure}
}

\newcommand{\suppunpairedtranslation}{
\begin{figure}
  \centering
  \renewcommand{\impatha}[1]{img/facelift_samples/portraits/##1}
  \renewcommand{\impathb}[1]{img/facelift_samples/animes/##1}
  \renewcommand{\impathc}[1]{img/facelift_samples/celebify/##1}
  \renewcommand{\imheight}{3.8em}
  \renewcommand{\imwidth}{3.8em}
  \setlength{\tabcolsep}{1pt}
  \begin{tabular}{c   c c c c c c c c}
    \toprule
    \scriptsize{photography} & 
    \includegraphics[width=\imwidth, align=c]{\impatha{lq-0000}} &
    \includegraphics[width=\imwidth, align=c]{\impatha{lq-0001}} &
    \includegraphics[width=\imwidth, align=c]{\impatha{lq-0002}} &
    \includegraphics[width=\imwidth, align=c]{\impatha{lq-0003}} &
    \includegraphics[width=\imwidth, align=c]{\impatha{lq-0004}} &
    \includegraphics[width=\imwidth, align=c]{\impatha{lq-0005}} &
    \includegraphics[width=\imwidth, align=c]{\impatha{lq-0006}} &
    \includegraphics[width=\imwidth, align=c]{\impatha{lq-0007}} \\
    
    \scriptsize{oil portrait} & 
    \includegraphics[width=\imwidth, align=c]{\impatha{hq-0000}} &
    \includegraphics[width=\imwidth, align=c]{\impatha{hq-0001}} &
    \includegraphics[width=\imwidth, align=c]{\impatha{hq-0002}} &
    \includegraphics[width=\imwidth, align=c]{\impatha{hq-0003}} &
    \includegraphics[width=\imwidth, align=c]{\impatha{hq-0004}} &
    \includegraphics[width=\imwidth, align=c]{\impatha{hq-0005}} &
    \includegraphics[width=\imwidth, align=c]{\impatha{hq-0006}} &
    \includegraphics[width=\imwidth, align=c]{\impatha{hq-0007}} \\
    \midrule

    \scriptsize{photography} &
    \includegraphics[width=\imwidth, align=c]{\impathb{lq-0000}} &
    \includegraphics[width=\imwidth, align=c]{\impathb{lq-0001}} &
    \includegraphics[width=\imwidth, align=c]{\impathb{lq-0002}} &
    \includegraphics[width=\imwidth, align=c]{\impathb{lq-0003}} &
    \includegraphics[width=\imwidth, align=c]{\impathb{lq-0008}} &
    \includegraphics[width=\imwidth, align=c]{\impathb{lq-0009}} &
    \includegraphics[width=\imwidth, align=c]{\impathb{lq-0010}} &
    \includegraphics[width=\imwidth, align=c]{\impathb{lq-0007}} \\
    
    \scriptsize{anime} & 
    \includegraphics[width=\imwidth, align=c]{\impathb{hq-0000}} &
    \includegraphics[width=\imwidth, align=c]{\impathb{hq-0001}} &
    \includegraphics[width=\imwidth, align=c]{\impathb{hq-0002}} &
    \includegraphics[width=\imwidth, align=c]{\impathb{hq-0003}} &
    \includegraphics[width=\imwidth, align=c]{\impathb{hq-0008}} &
    \includegraphics[width=\imwidth, align=c]{\impathb{hq-0009}} &
    \includegraphics[width=\imwidth, align=c]{\impathb{hq-0010}} &
    \includegraphics[width=\imwidth, align=c]{\impathb{hq-0007}} \\
	\midrule
    \bottomrule
  \end{tabular}
   \caption{Additional examples for unpaired translation of Oil Portraits to FFHQ/CelebA-HQ and Anime to FFHQ/CelebA-HQ. Here, we show samples where the \emph{same} $\modelinv$ is projected onto the respective dataset.}
  \label{fig:suppunpaired}
\end{figure}
}

\newcommand{\suppanimalswap}{
\begin{figure}
  \centering
  \renewcommand{\impath}[1]{img/suppanimalswap/##1}
  \renewcommand{\imwidth}{0.13\textwidth}
  \renewcommand{\imheight}{3.8em}
  \setlength{\tabcolsep}{2pt}
  \begin{tabular}{cc}
    \footnotesize{$\oneenc(\x)$\;\;\;}\rotatebox{90}{\scriptsize{exemplar $\y$}}&
    \includegraphics[height=\imheight]{\impath{exemplar_000000}}%
    \includegraphics[height=\imheight]{\impath{exemplar_000001}}%
    \includegraphics[height=\imheight]{\impath{exemplar_000002}}%
    \includegraphics[height=\imheight]{\impath{exemplar_000003}}%
    \includegraphics[height=\imheight]{\impath{exemplar_000004}}%
    \includegraphics[height=\imheight]{\impath{exemplar_000005}}%
    \includegraphics[height=\imheight]{\impath{exemplar_000006}}%
    \includegraphics[height=\imheight]{\impath{exemplar_000007}} \\

\toprule

          \includegraphics[height=\imheight, align=c]{\impath{ctxt_000000}} &     
    \includegraphics[height=\imheight, align=c]{\impath{x_ss_cross_000000}} \\

          \includegraphics[height=\imheight, align=c]{\impath{ctxt_000001}} &     
    \includegraphics[height=\imheight, align=c]{\impath{x_ss_cross_000001}} \\

          \includegraphics[height=\imheight, align=c]{\impath{ctxt_000002}} &     
    \includegraphics[height=\imheight, align=c]{\impath{x_ss_cross_000002}} \\

          \includegraphics[height=\imheight, align=c]{\impath{ctxt_000003}} &     
    \includegraphics[height=\imheight, align=c]{\impath{x_ss_cross_000003}} \\

          \includegraphics[height=\imheight, align=c]{\impath{ctxt_000004}} &     
    \includegraphics[height=\imheight, align=c]{\impath{x_ss_cross_000004}} \\

          \includegraphics[height=\imheight, align=c]{\impath{ctxt_000005}} &     
    \includegraphics[height=\imheight, align=c]{\impath{x_ss_cross_000005}} \\

          \includegraphics[height=\imheight, align=c]{\impath{ctxt_000006}} &     
    \includegraphics[height=\imheight, align=c]{\impath{x_ss_cross_000006}} \\

          \includegraphics[height=\imheight, align=c]{\impath{ctxt_000007}} &     
    \includegraphics[height=\imheight, align=c]{\impath{x_ss_cross_000007}} \\

	\bottomrule
  \end{tabular}
   \caption{Additional examples for exemplar-guided image-to-image translation as in Fig.~\ref{fig:animalswap}.}
  \label{fig:suppanimalswap}
\end{figure}
}

\newcommand{\suppunsupswap}{
\begin{figure}
  \centering
  \renewcommand{\impath}[1]{img/suppunsupswap/##1}
  \renewcommand{\imwidth}{0.13\textwidth}
  \renewcommand{\imheight}{3.8em}
  \setlength{\tabcolsep}{2pt}
  \begin{tabular}{cc}
    \footnotesize{\hspace{1.5em}$\x$\;\;\;\;\;}\rotatebox{90}{\hspace{1.5em}$\y$}&
    \includegraphics[height=\imheight]{\impath{exemplar_000007}}%
    \includegraphics[height=\imheight]{\impath{exemplar_000008}}%
    \includegraphics[height=\imheight]{\impath{exemplar_000009}}%
    \includegraphics[height=\imheight]{\impath{exemplar_000010}}%
    \includegraphics[height=\imheight]{\impath{exemplar_000011}}%
    \includegraphics[height=\imheight]{\impath{exemplar_000012}}%
    \includegraphics[height=\imheight]{\impath{exemplar_000013}} \\

\toprule

          \includegraphics[height=\imheight, align=c]{\impath{ctxt_000007}} &     
    \includegraphics[height=\imheight, align=c]{\impath{x_ss_cross_000007}} \\

          \includegraphics[height=\imheight, align=c]{\impath{ctxt_000008}} &     
    \includegraphics[height=\imheight, align=c]{\impath{x_ss_cross_000008}} \\

          \includegraphics[height=\imheight, align=c]{\impath{ctxt_000009}} &     
    \includegraphics[height=\imheight, align=c]{\impath{x_ss_cross_000009}} \\

          \includegraphics[height=\imheight, align=c]{\impath{ctxt_000010}} &     
    \includegraphics[height=\imheight, align=c]{\impath{x_ss_cross_000010}} \\

          \includegraphics[height=\imheight, align=c]{\impath{ctxt_000011}} &     
    \includegraphics[height=\imheight, align=c]{\impath{x_ss_cross_000011}} \\

          \includegraphics[height=\imheight, align=c]{\impath{ctxt_000012}} &     
    \includegraphics[height=\imheight, align=c]{\impath{x_ss_cross_000012}} \\

          \includegraphics[height=\imheight, align=c]{\impath{ctxt_000013}} &     
    \includegraphics[height=\imheight, align=c]{\impath{x_ss_cross_000013}} \\

	\bottomrule
  \end{tabular}
   \caption{Unsupervised disentangling of shape and appearance. Training our
   approach on
   synthetically deformed images, $\condinn$ learns to extract a
   disentangled shape representation $\modelinv$ from $\y$, which can be
   recombined with arbitrary appearances obtained from $\x$. See also
   Sec.~\ref{supp:unsupervised}.}
  \label{supp:unsupswap}
\end{figure}
}

\newcommand{\unsupswap}{
  \centering
  \renewcommand{\impath}[1]{img/suppunsupswapcrop/##1}
  \renewcommand{\imwidth}{0.13\textwidth}
  \renewcommand{\imheight}{3.8em}
  \setlength{\tabcolsep}{2pt}
  \begin{tabular}{cc}
    \footnotesize{$\oneenc(\x)$\;\;\;}\rotatebox{90}{\scriptsize{exemplar $\y$}}&
    \includegraphics[height=\imheight]{\impath{exemplar_000007}}%
    \includegraphics[height=\imheight]{\impath{exemplar_000008}}%
    \includegraphics[height=\imheight]{\impath{exemplar_000009}}%
    \includegraphics[height=\imheight]{\impath{exemplar_000010}} \\%

\toprule

          \includegraphics[height=\imheight, align=c]{\impath{ctxt_000007}} &     
    \includegraphics[height=\imheight, align=c]{\impath{x_ss_cross_000007}} \\

          \includegraphics[height=\imheight, align=c]{\impath{ctxt_000008}} &     
    \includegraphics[height=\imheight, align=c]{\impath{x_ss_cross_000008}} \\

          \includegraphics[height=\imheight, align=c]{\impath{ctxt_000010}} &     
    \includegraphics[height=\imheight, align=c]{\impath{x_ss_cross_000010}} \\

	\bottomrule
  \end{tabular}
   \caption{Unsupervised shape and appearance disentangling.}
  \label{fig:unsupswap}
}

\newcommand{\suppanimallayers}{
\begin{figure}[tb]
  \renewcommand{\impath}[1]{img/suppanimallayers/##1}
  \renewcommand{\imwidth}{0.055\textwidth}
  \renewcommand{\imwidthmult}{0.275\textwidth}
  \setlength{\tabcolsep}{2pt}
  \centering
  \begin{tabular}{c c  c  c  c}
    \footnotesize{$\x$} & method & \footnotesize{early layer: $\twodec(\reptwo)$} &
    \footnotesize{middle layer: $\twodec(\reptwo)$} & \footnotesize{last
    layer of $\modelone$: $\twodec(\reptwo)$} \\
    \toprule
    \multirow{2}{*}{\includegraphics[width=\imwidth, align=c]{\impath{ours/lm1/exemplar_000025}}} &
            our &
    \includegraphics[width=\imwidthmult, align=c]{\impath{ours/lm5/x_ss_sample_000025}} &     
    \includegraphics[width=\imwidthmult, align=c]{\impath{ours/lm2/x_ss_sample_000025}} &     
    \includegraphics[width=\imwidthmult, align=c]{\impath{ours/lm1/x_ss_sample_000025}} \\

 	         &         
            MLP &
    \includegraphics[width=\imwidthmult, align=c]{\impath{mlp/lm5/x_ss_sample_000025}} &     
    \includegraphics[width=\imwidthmult, align=c]{\impath{mlp/lm2/x_ss_sample_000025}} &     
    \includegraphics[width=\imwidthmult, align=c]{\impath{mlp/lm1/x_ss_sample_000025}} \\

    \midrule

    \multirow{2}{*}{\includegraphics[width=\imwidth, align=c]{\impath{ours/lm1/exemplar_000026}}} &
            our &
    \includegraphics[width=\imwidthmult, align=c]{\impath{ours/lm5/x_ss_sample_000026}} &     
    \includegraphics[width=\imwidthmult, align=c]{\impath{ours/lm2/x_ss_sample_000026}} &     
    \includegraphics[width=\imwidthmult, align=c]{\impath{ours/lm1/x_ss_sample_000026}} \\

    &
            MLP &
    \includegraphics[width=\imwidthmult, align=c]{\impath{mlp/lm5/x_ss_sample_000026}} &     
    \includegraphics[width=\imwidthmult, align=c]{\impath{mlp/lm2/x_ss_sample_000026}} &     
    \includegraphics[width=\imwidthmult, align=c]{\impath{mlp/lm1/x_ss_sample_000026}} \\

    \midrule

    \multirow{2}{*}{\includegraphics[width=\imwidth, align=c]{\impath{ours/lm1/exemplar_000027}}} &
            our &
    \includegraphics[width=\imwidthmult, align=c]{\impath{ours/lm5/x_ss_sample_000027}} &     
    \includegraphics[width=\imwidthmult, align=c]{\impath{ours/lm2/x_ss_sample_000027}} &     
    \includegraphics[width=\imwidthmult, align=c]{\impath{ours/lm1/x_ss_sample_000027}} \\

 	         &         
            MLP &
    \includegraphics[width=\imwidthmult, align=c]{\impath{mlp/lm5/x_ss_sample_000027}} &     
    \includegraphics[width=\imwidthmult, align=c]{\impath{mlp/lm2/x_ss_sample_000027}} &     
    \includegraphics[width=\imwidthmult, align=c]{\impath{mlp/lm1/x_ss_sample_000027}} \\

    \midrule

    \multirow{2}{*}{\includegraphics[width=\imwidth, align=c]{\impath{ours/lm1/exemplar_000028}}} &
            our &
    \includegraphics[width=\imwidthmult, align=c]{\impath{ours/lm5/x_ss_sample_000028}} &     
    \includegraphics[width=\imwidthmult, align=c]{\impath{ours/lm2/x_ss_sample_000028}} &     
    \includegraphics[width=\imwidthmult, align=c]{\impath{ours/lm1/x_ss_sample_000028}} \\

 	         &         
            MLP &
    \includegraphics[width=\imwidthmult, align=c]{\impath{mlp/lm5/x_ss_sample_000028}} &     
    \includegraphics[width=\imwidthmult, align=c]{\impath{mlp/lm2/x_ss_sample_000028}} &     
    \includegraphics[width=\imwidthmult, align=c]{\impath{mlp/lm1/x_ss_sample_000028}} \\

    \midrule

    \multirow{2}{*}{\includegraphics[width=\imwidth, align=c]{\impath{ours/lm1/exemplar_000029}}} &
            our &
    \includegraphics[width=\imwidthmult, align=c]{\impath{ours/lm5/x_ss_sample_000029}} &     
    \includegraphics[width=\imwidthmult, align=c]{\impath{ours/lm2/x_ss_sample_000029}} &     
    \includegraphics[width=\imwidthmult, align=c]{\impath{ours/lm1/x_ss_sample_000029}} \\

 	         &         
            MLP &
    \includegraphics[width=\imwidthmult, align=c]{\impath{mlp/lm5/x_ss_sample_000029}} &     
    \includegraphics[width=\imwidthmult, align=c]{\impath{mlp/lm2/x_ss_sample_000029}} &     
    \includegraphics[width=\imwidthmult, align=c]{\impath{mlp/lm1/x_ss_sample_000029}} \\

    \midrule

    \multirow{2}{*}{FID} &
            our &
    $34.0 \pm 0.1$ &     
    $23.4 \pm 0.7$ &     
    $27.6 \pm 0.1$ \\

 	         &         
            MLP &
    $24.2$ &     
    $25.6$ &     
    $264.0$ \\

    \bottomrule  
  \end{tabular}    
  \caption{Model diagnosis compared to a MLP for the translation. Synthesized samples and FID scores demonstrate that a direct translation with a multilayer perceptron (MLP) does not capture the ambiguities of the translation process and can thus only produce a mean image. In contrast, our cINN correctly captures the variability and produces coherent outputs.}
  \label{fig:suppanimallayers}
\end{figure}
}

\newcommand{\supplandscapesamples}{
\begin{figure}
  \centering
  \renewcommand{\impath}[1]{img/supplandscapesamples/##1}
  \renewcommand{\imwidth}{0.13\textwidth}
  \renewcommand{\imheight}{3.8em}
  \setlength{\tabcolsep}{2pt}
  \begin{tabular}{cc}
    \footnotesize{$\oneenc(\x)$} & translating $\oneenc(\x)$ onto target domain
    of AE $\modeltwo$ with different samples $\modelinv \sim q(\modelinv)$ \\

\toprule

          \includegraphics[height=\imheight, align=c]{\impath{ctxt_000000}} &     
    \includegraphics[height=\imheight, align=c]{\impath{x_ss_sample_000000}} \\
    \midrule

          \includegraphics[height=\imheight, align=c]{\impath{ctxt_000001}} &     
    \includegraphics[height=\imheight, align=c]{\impath{x_ss_sample_000001}} \\
    \midrule

          \includegraphics[height=\imheight, align=c]{\impath{ctxt_000002}} &     
    \includegraphics[height=\imheight, align=c]{\impath{x_ss_sample_000002}} \\
    \midrule

          \includegraphics[height=\imheight, align=c]{\impath{ctxt_000004}} &     
    \includegraphics[height=\imheight, align=c]{\impath{x_ss_sample_000004}} \\
    \midrule

          \includegraphics[height=\imheight, align=c]{\impath{ctxt_000005}} &     
    \includegraphics[height=\imheight, align=c]{\impath{x_ss_sample_000005}} \\
    \midrule

          \includegraphics[height=\imheight, align=c]{\impath{ctxt_000006}} &     
    \includegraphics[height=\imheight, align=c]{\impath{x_ss_sample_000006}} \\
    \midrule

          \includegraphics[height=\imheight, align=c]{\impath{ctxt_000007}} &     
    \includegraphics[height=\imheight, align=c]{\impath{x_ss_sample_000007}} \\

	\bottomrule
  \end{tabular}
   \caption{Additional \emph{Landscape} samples, obtained by translation of the argmaxed logits (\ie the segmentation output) of the segmentation model from Sec.~\ref{subsec:reuse},~\ref{subsec:imgmod} into the space of our autoencoder $\modeltwo$, see Sec.~\ref{subsec:reuse},~\ref{subsec:imgmod}. The synthesized examples demonstrate that our approach is able to generate diverse and realistic images from a given label map or through a segmentation model.}
  \label{fig:supplandscapesamples}
\end{figure}
}

\begin{document}
\maketitle
\setlength{\belowcaptionskip}{-1pt}
\begin{abstract}
Given the ever-increasing computational costs of modern machine learning models, we need to find new ways to reuse such expert models and thus tap into the resources that have been invested in their creation. Recent work suggests that the power of these massive models is captured by the representations they learn. Therefore, 
we seek a model that can relate between different existing representations and propose to solve this task with a conditionally invertible network. This network
  demonstrates its capability by
  (i) providing generic transfer between diverse domains, (ii) enabling
  controlled content synthesis by allowing modification in other domains,
  and (iii) facilitating diagnosis of existing representations by
  translating them into interpretable domains such as images. Our domain transfer
  network can translate between fixed representations without having to learn
  or finetune them. This allows users to utilize various existing
  domain-specific expert models from the literature that had been trained
  with extensive computational resources.
  Experiments on diverse
  conditional image synthesis tasks, competitive image modification results
  and experiments on image-to-image and text-to-image generation
  demonstrate the generic applicability of our approach. For example, we
  translate between BERT and BigGAN, state-of-the-art text and image models to
  provide text-to-image generation, which neither of both experts can perform
  on their own.
\end{abstract}

\section{Introduction}
One of the key features of intelligence is the ability to combine and transfer
information between diverse domains and modalities \cite{crick1990towards, von1995binding, singer2001consciousness, treisman1996binding, seymour2009coding}.
In contrast, artificial intelligence research has made great progress in learning powerful
representations for \emph{individual} domains \cite{Han_2019,
oord2018representation, donahue2019large, oord2016wavenet, devlin2018bert,
brown2020language} that can even achieve superhuman performance on confined tasks
such as traffic sign recognition \cite{cirecsan2011committee,
cirecsan2012multi}, image classification \cite{he2016deep} or question
answering \cite{devlin2018bert}.
However, learning representations for different domains that also allow a
domain-to-domain transfer of information between them is significantly
more challenging \cite{Baltrusaitis_2019}: There is a trade-off between the
expressiveness of individual domain representations and their compatibility to
another to support transfer.  While for limited training data multimodal
learning has successfully trained representations for different domains
together \cite{srivastava2012multimodal,wang2015deep}, the overall most
powerful domain-specific representations typically result from training huge
models specifically for \emph{individual} challenging domains using massive
amounts of training data and computational resources, \eg
\cite{donahue2019large,oord2016wavenet,brown2020language}.
\berttobigfirstpage
With the dawn of even more massive models like the recently introduced GPT-3
\cite{brown2020language}, where training on only a single domain already
demands most of the available resources, we must find new, creative ways to
make use of these powerful models, which none but the largest institutions
can afford to train and experiment with, and thereby utilize the huge amount of
resources and knowledge which are distilled into the model's representations---in other words, 
we have to find ways to cope with "The Bitter Lesson" \cite{bitterlesson}.

Consequently, we seek a model for generic domain-to-domain transfer between
arbitrary fixed representations that come from highly complex, off-the-shelf,
state-of-the-art models
and we learn a domain translation that does not alter or retrain the individual
representations but retains the full capabilities of original expert models.
This stands in contrast to current influential domain transfer approaches
\cite{liu2019few,park2019semantic,zhang2020cross} that require learning or
finetuning existing domain representations to facilitate transfer between them. 

Since different domains are typically not isomorphic to another, \ie
translations between them are not uniquely determined, the domain translation
between fixed domain representations requires learning the corresponding
ambiguities. For example, there are many images which correspond to the same
textual description and vice versa.
To faithfully translate between domains, we employ a conditional invertible neural network (cINN) that also explicitly captures these transfer uncertainties. The INN conditionally learns a unique translation of one domain representation together with its complementary residual onto another. This generic network-to-network translation between arbitrary models can efficiently transfer between diverse state-of-the-art models such as transformer-based natural language model BERT
\cite{devlin2018bert} and a BigGAN \cite{brock2018large} for image synthesis to achieve competitive text-to-image translation, see Fig.~\ref{fig:berttobig}.

To summarize, our contributions are as follows: We (i) provide a generic
approach that allows to translate between fixed off-the-shelf model
representations, (ii) learns the inherent ambiguity of the domain translation,
which facilitates content creation and model diagnostics, and (iii) enables
compelling performance on various different domain transfer problems. We
(iv) make transfer between domains and datasets computationally affordable,
since our method does not require any gradient computations on the expert
models but can directly utilize existing representations.

\section{Related Work}
\label{sec:background}
The majority of approaches for deep-learning-based domain-to-domain translation
are based on generative models and therefore rely on Variational Autoencoders
(VAEs) \cite{VAE,VAE2}, Generative Adversarial Networks (GANs) \cite{gan},
autoregressive models \cite{ar}, or normalizing flows
\cite{papamakarios2019normalizing} obtained with invertible neural networks
(INNs) \cite{dinh2014nice,dinh2016density}. Generative models transform samples
from a simple base distribution, mainly a standard normal or a uniform
distribution, to a complex target distribution, \eg the distribution of (a
subset of) natural images.

Sampling the base distribution then leads to the generation of novel content.
Recent works \cite{xiao2019generative,esser2020invertible} also utilize INNs
to transform the latent distribution of an autoencoder to the base
distribution.
A simple structure of the base distribution allows rudimentary control over the
generative process in the form of vector arithmetic applied to samples
\cite{radford2015unsupervised,NIPS2015_5845,shen2019interpreting,goetschalckx2019ganalyze},
but more generally, providing control over the generated content is formulated
as conditional image synthesis.
In its most basic form, conditional image synthesis is achieved by generative
models which, in addition to a sample from the base distribution, take class
labels \cite{mirza2014conditional,kingma2014semi} or attributes
\cite{he2019attgan} into account. More complex conditioning information are
considered in \cite{zhang2017stackgan,reed2016generative}, where textual
descriptions provide more fine-grained control over the generative process.  A
wide range of approaches can be characterized as image-to-image translations
where both the generated content and the conditioning information is given by
images. Examples for conditioning images include grayscale images
\cite{zhang2016colorful}, low resolution images \cite{ledig2017photo}, edge
images \cite{isola2017image}, segmentation maps
\cite{park2019semantic,chen2017photographic} or heatmaps of keypoints
\cite{Esser_2018,ma2017disentangled}.  Many of these approaches build upon
\cite{isola2017image}, which introduced a unified approach for image-to-image
translation. We take this unification one step further and provide an approach
for a wide range of conditional content creation, including class labels,
attributes, text and images as conditioning. In the case of image conditioning,
our approach can be trained either with aligned image pairs as in
\cite{isola2017image,chen2017photographic,park2019semantic} or with unaligned
image pairs as in
\cite{Zhu_2017,Lee_2018,choi2019stargan,Huang_2018,esser2019unsupervised}.
\modelneurips
While many works on generative models focus on relatively simple datasets
containing little variations, \eg CelebA \cite{liu2015faceattributes}
containing only aligned images of faces, \cite{brock2018large,
donahue2019large} demonstrated the possibility to apply these models to
large-scale datasets such as ImageNet \cite{deng2009imagenet}. However, such
experiments require a computational effort which is typically far out of reach
for individuals. Moreover, the need to retrain large models for experimentation
hinders rapid prototyping of new ideas and thus slows down progress. Making use
of pre-trained neural networks can significantly reduce the computational
budget and training time. For discriminative tasks, the ability to effectively
reuse pre-trained neural networks has long been recognized
\cite{Razavian_2014,donahue2013decaf,yosinski2014transferable}. For generative
tasks, however, there are less works that aim to reuse pre-trained networks
efficiently. Features obtained from pre-trained classifier networks are used to
derive style and content losses for style transfer algorithms
\cite{Gatys_2016}, and they have been demonstrated to measure perceptual
similarity between images significantly better than pixelwise distances
\cite{mahendran2015understanding,zhang2018unreasonable}.
\cite{yosinski2015understanding,mahendran2016visualizing} find images which
maximally activate neurons of pre-trained networks and
\cite{santurkar2019image} shows that improved synthesis results are obtained
with adversarially robust classifiers. Instead of directly searching over
images, \cite{nguyen2016synthesizing} uses a pre-trained generator network of
\cite{dosovitskiy2016generating}, where it was used to reconstruct images from
feature representations.  However, these approaches are limited to neuron
activation problems, rely on per-example optimization problems, which makes
synthesis slow, and do not take into account the probabilistic nature of the
conditional synthesis task, where a single conditioning corresponds to multiple
outputs. To address this, \cite{nash2018inverting} learns an autoregressive
model conditioned on specific layers of pre-trained models and
\cite{shocher2020semantic} a GAN based decoder conditioned on a feature pyramid
of a pre-trained classifier. In contrast, our approach efficiently utilizes
pre-trained models, both for conditioning as well as for image-synthesis, such
that their combination provides new generative capabilities for content
creation through conditional sampling, without requiring the pre-trained models
to be aware of these emerging capabilities.

\section{Approach}
\label{sec:approach}

Our goal is to learn relationships and transfer between representations of
different domains obtained from off-the-shelf models, see
Fig.~\ref{fig:modelnips}.
To be generally applicable to complex state-of-the-art representations, we only
assume the availability of already trained models, but no practical access to
their training procedure due to their complexity \cite{devlin2018bert} or
missing components (\eg a discriminator, which was not released for
\cite{donahue2019large}).
Let
$\domainone$ and $\domaintwo$ be two domains we want to transfer between.
Moreover, $\modelone(\x)$ denotes an expert model that has been trained to map
$\x \in \domainone$ onto desired outputs, \eg class labels in case of
classification tasks, or synthesized images for generative image models. To
solve its task, a neural network $\modelone$ has learned a latent
representation $\repone = \oneenc(\x)$ of domain $\domainone$ in some
intermediate layer, so that subsequent layers $\onedec$ can then solve the task as
$\modelone(\x) = \onedec(\oneenc(\x))$. For $\y \in \domaintwo$ let
$\modeltwo(\y) = \twodec(\twoenc(\y))$ be another, totally different model that
provides a feature representation vector $\reptwo = \twoenc(\y)$.

In general, we cannot expect a translation from $\x$ to $\y$ to be unique,
since two arbitrary domains and their representations are not necessarily
isomorphic.
For example, a textual description $\x$ of an image $\y$ usually leaves many
details open and the same holds in the opposite direction, since many
different textual descriptions are conceivable for the same image by focusing
on different aspects. This
implies a non-unique mapping from $\repone$ to $\reptwo$.
Moreover, much of the power of model $\modelone$ trained for a specific task
stems from its ability to ignore task-irrelevant properties of $\x$.
The invariances of $\repone$ with respect to $\reptwo$ further increase the
ambiguity of the domain translation. Obtaining a plausible $\reptwo$ for a given
$\repone$ is therefore best described probabilistically as sampling from
$p(\reptwo \vert \repone)$.
Our goal is to model this process with a translation function $\condinn$. Thus, we must
introduce a residual $\modelinv$,
such that for a given $\repone$, $\modelinv$ uniquely determines $\reptwo$
resulting in the
translation function $\condinn$:
\begin{equation}
  \reptwo = \condinn(\modelinv \vert \repone)
\end{equation}

\newcommand{\tmpinn}{u}
\paragraph{Learning a Domain Translation $\condinn$:}
How can we estimate $\modelinv$? $\modelinv$ must capture all information of
$\reptwo$ not represented in $\repone$, but
no information that is already represented in $\repone$.
Hence, to infer $\modelinv$, we must take into account both
$\reptwo$, to extract information, and $\repone$, to discard information.
The unique determination of $\reptwo$ from $\modelinv$ for a given $\repone$ implies
the
existence of the inverse of $\condinn$, when considered as a function of
$\modelinv$. Thus for every $\repone$, the inverse $\condinn^{-1}(\cdot \vert
\repone)$ of $\condinn(\cdot \vert \repone)$ exists,
\begin{equation}
  \label{eq:invertible}
  \modelinv = \condinn^{-1}(\reptwo \vert \repone) .
\end{equation}

This structure of $\condinn$ is most naturally represented by a conditionally
invertible neural network (cINN), for which $\condinn^{-1}$ can be explicitly
computed, and which we build from affine coupling \cite{dinh2016density},
actnorm \cite{kingma2018glow} and shuffling layers, see Sec.~\ref{subsec:cinnarch}. It then remains to derive a learning task which ensures that information of $\repone$ is discarded in
$\modelinv$. To formalize this goal, we consider training pairs
$\{(\x, \y)\} \subset \domainone \times \domaintwo$ and their corresponding features
$\{(\repone, \reptwo)\}$ as samples from their joint distribution $p(\repone,
\reptwo)$. $\modelinv$ can then be considered as a random variable via the
process
\begin{equation}
  \label{eq:induced}
  \modelinv = \condinn^{-1}(\reptwo \vert \repone), \quad \text{with }\repone, \reptwo \sim
  p(\repone, \reptwo).
\end{equation}
Then $\modelinv$ discards all information of $\repone$ if
$\modelinv$ and $\repone$ are independent. To achieve this independence, we
minimize the distance between the distribution $p(\modelinv \vert \repone)$
induced by $\condinn$ via Eq. ~\eqref{eq:induced} and some prior distribution
$q(\modelinv)$. The latter can be chosen arbitrarily as long as it is
independent of $\repone$, its density can be evaluated and samples can be
drawn. In practice we use a standard normal distribution.
Using the invertibility of $\condinn$, we can then explicitly calculate the
Kullback-Leibler divergence between $p(\modelinv \vert \repone)$ and
$q(\modelinv)$ averaged over $\repone$ (see Sec.~\ref{suppsec:objective} for the derivation):
\begin{align}
  \label{eq:loss}
  \expect_{\repone} \KL(p(\modelinv \vert \repone) \vert q(v)) =
  &\expect_{\reptwo, \repone}
  \left\{
    -\log q(\condinn^{-1}(\reptwo \vert \repone))
    - \vert \det J_{\condinn^{-1}}(\reptwo \vert \repone) \vert
  \right\}
  -H(\reptwo \vert \repone).
\end{align}
Here, $\det J_{\condinn^{-1}}$ denotes the determinant of the Jacobian of $\condinn^{-1}$ and $H$ is the (constant) data entropy. If $\condinn$ minimizes Eq.~\eqref{eq:loss}, we have $p(\modelinv \vert
\repone) = q(v)$, such that the desired independence is achieved. Moreover, we
can now simply achieve the
original goal of sampling from $p(\reptwo \vert \repone)$ by
translating from $\repone$ to $\reptwo = \condinn(\modelinv \vert \repone)$
with $\modelinv$ sampled from $q(\modelinv)$, which properly models
the inherent ambiguity. %

\paragraph{Interpretation as Information Bottleneck:}
One of the main goals of minimizing Eq.~\eqref{eq:loss} is the independence of
$\modelinv$ and $\repone$. While it is clear that this independence is achieved by a
minimizer, Eq.~\eqref{eq:loss} is also an upper bound on the mutual information
$I(\modelinv, \repone)$ between $\modelinv$ and $\repone$. Thus, its
minimization works directly towards the goal of independence. Indeed, following
\cite{alex2016deep}, we have
\begin{equation}
  I(\modelinv, \repone) = \int_{\modelinv,\repone} p(\modelinv, \repone) \log \frac{p(\modelinv,
  \repone)}{p(\modelinv) p(\repone)} = \int_{\modelinv, \repone} p(\modelinv, \repone) \log
  p(\modelinv \vert \repone) - \int_{\modelinv} p(\modelinv) \log p(\modelinv)
\end{equation}
Positivity of the KL divergence implies $\int p(\modelinv) \log p(\modelinv)
\ge \int p(\modelinv) \log q(\modelinv)$, such that
\begin{equation}
  I(\modelinv, \repone) \le \int_{\modelinv, \repone} p(\modelinv, \repone) \log \frac{p(\modelinv
  \vert \repone)}{q(\modelinv)} = \expect_\repone \KL(p(\modelinv \vert
  \repone) \vert q(\modelinv)) = \text{Eq.~\eqref{eq:loss}}
  \label{eq:infobound}
\end{equation}
In contrast to the deep variational information bottleneck \cite{alex2016deep},
our use of a cINN has the advantage that it does not require the hyperparameter
$\beta$ to balance the independence of $\modelinv$ and $\repone$ against their
ability to reconstruct $\reptwo$. The cINN guarantees perfect reconstruction
abilities of $\reptwo$ due to its invertible architecture and it thus suffices
to minimize $I(\modelinv, \repone)$ on its own.

\paragraph{Domain Transfer Between Fixed Models:}
At inference time, we obtain translated samples $\reptwo$ for given $\repone$
by sampling from the residual space $\modelinv$ given $\repone$ and then
applying $\condinn$,
    \begin{equation}
      \reptwo \sim p(\reptwo \vert \repone) %
      \quad \iff \quad %
      \modelinv \sim q(\modelinv) %
      \label{eq:sample}
      ,\;
      \reptwo= \condinn(\modelinv \vert \repone) .
    \end{equation}
After training our domain translator, transfer between $\domainone$ and $\domaintwo$ is thus
achieved by the following steps: (i) sample $\x$ from $p(\x)$, (ii) encode $\x$
into the latent space $\repone = \oneenc(\x)$ of expert model $\modelone$,
(iii) sample a residual $\modelinv$ from the prior $q(\modelinv)$, (iv)
conditionally transform $\reptwo = \condinn(\modelinv \vert \repone)$, and (v)
decode $\reptwo$ into the domain $\domaintwo$ of the second expert model: $\y =
\twodec(\reptwo)$.

Note that this approach has multiple advantages: (i) hidden representations
usually have lower dimensionality than $\x$, which makes transfer between
arbitrary complex domains affordable, (ii) the cINN $\condinn$ can
be trained by minimizing the negative log-likelihood, independent of the
domains $\domainone$ and $\domaintwo$, and (iii) the approach does not require
to take any gradients \wrt the expert models $\modelone$ and $\modeltwo$.

\section{Experiments}
We investigate the wide applicability of our approach by performing experiments
with multiple domains, datasets and models: (1) text-to-image translation by
combination of BigGAN and BERT, (2) exploration of the reusability of a fixed
autoencoder combined with multiple models including a ResNet-50 classifier and a
DeepLabV2 \cite{chen2017deeplab} segmentation model for various image-to-image
translation tasks and diagnostic insights into the respective models, and (3)
comparison to existing methods for image modification and applications in
exemplar-guided and unpaired image translation tasks.
As our method does not require gradients \wrt the models $\modelone$
and $\modeltwo$, training of the cINN can be conducted on a single Titan X GPU.

\berttobiginception
\subsection{Translation to BigGAN}
\label{subsec:transtobig}
This section is dedicated to the task of using a popular but computationally expensive to train expert model as an image generator: BigGAN
\cite{brock2018large}, achieving state-of-the-art FID scores \cite{heusel2017gans} on the ImageNet dataset. As most GAN frameworks in general and BigGAN in particular do not
include an explicit encoder into a latent space, we aim to provide an encoding from an arbitrary domain by using an appropriate expert model $\modelone$. 
Aiming at the reusability of a fixed BigGAN $\modeltwo$, and given the hidden
representation $\repone = \oneenc(\x)$ of the expert model $\modelone=\onedec
\circ \oneenc$, we want to find a mapping between $\repone$ and the latent
space $\reptwo$ of BigGAN's generator $\twodec$, where, in accordance with
Fig.~\ref{fig:modelnips}, $\twoenc \equiv \id$ and $\modeltwo = \twodec$.
Technical details regarding the training of our cINN can be found in
Sec.~\ref{subsec:berttobigdetails}. \newline
\vspace{-0.65cm}
\paragraph{BERT-to-BigGAN Translation:}
The emergence of transformer-based networks \cite{vaswani2017attention} has led to an immense leap in the field of natural language processing, where a popular model is the so-called BERT model. Here, we make use of a variant of the original model, which modifies BERT such that it produces a latent space in which input sentences can be compared for similarity via the cosine-distance measure \cite{reimers-2019-sentence-bert}. We aim to combine this representational power with the synthesis capabilities of BigGAN and thus train our model $\condinn$ to map from the language representations $\repone = \oneenc(\x)$ into the latent space $\reptwo$ of BigGAN's generator as described above; hence $\modelone = \oneenc$ and $\onedec = \id$. During training, access to textual descriptions is obtained by using a captioning model as in \cite{xu2015show}, trained on the COCO \cite{lin2014microsoft} dataset. In a nutshell, at training time, we sample $\reptwo$, produce a corresponding image $\twodec(\reptwo)$, utilize \cite{xu2015show} to produce a text-caption $\x$ describing the image and subsequently produce a sentence representation $\repone = \oneenc(\x)$ which we use to minimize the overall objective Eq.~\eqref{eq:loss}. 
Results can be found in Fig.~\ref{fig:berttobig} and
Tab.~\ref{tab:bertinception}. Our model captures both fine-grained and coarse
descriptions (\eg blue bird vs. yellow bird; school bus vs. pizza) and is able
to synthesize images with highly different content, based on given textual
inputs $\x$. Although not being trained on the COCO images,
Tab.~\ref{tab:bertinception} shows that our model is highly competitive and
on-par with the state-of-the art in terms of Inception
\cite{salimans2016improved} and FID \cite{heusel2017gans} scores where
available.

\subsection{Repurposing a single target generator for different
source domain models}
\label{subsec:reuse}
Here, we train the cINN $\condinn$ conditioned on hidden representations of
networks such as classifiers and segmentation models, and thereby show that
standard classifiers on arbitrary source domains can drive the same generator
to create content by transfer.
Refering to
Fig.~\ref{fig:modelnips}, this means that $\modelone$ is represented by a
classifier/segmentation model, whereas $\twodec$ is a decoder of an
autoencoder that is pretrained on a dataset of interest. Furthermore, we
evaluate the ability of our approach to combine a single, powerful domain
expert (the autoencoder) with different source models to solve a variety of
image-to-image translation tasks.
\animaltable
The autoencoder is trained on a combination of all carnivorous animal classes
in ImageNet and images of the \emph{AwA2} dataset \cite{xian2018zero}, split
into 211306 training images and 10000 testing images, which we call the
\emph{Animals} dataset. The details regarding architecture and training of this
autoencoder are provided in Sec.~\ref{suppsec:aearch}.

\paragraph{Image-to-Image Translation:}
In Fig.~\ref{fig:animaltable}, we investigate the translation from different
source domain models $\oneenc$ onto the same generator $\twodec$
using our cINN $\condinn$.
In Fig.~\ref{tab:segtoimgargmax}, $\modelone$ is a segmentation network trained
on COCOStuff, and $\oneenc = \modelone$, \ie $\repone$ is given
by the final segmentation output of the network. This case corresponds to a
translation from segmentation masks to images and we observe that our approach
can successfully fuse the segmentation model with the autoencoder to obtain a
wide variety of generated image samples corresponding to a given segmentation mask.
Fig.~\ref{tab:segtoimglogits} uses the same segmentation network for
$\modelone$, but this time, $\oneenc$ are the logit predictions of the
network (visualized by a projection to RGB values). The diversity of
generated samples is greatly reduced compared to Fig.~\ref{tab:segtoimgargmax},
which indicates that logits still contain a lot of information which are not
strictly required for segmentation, \eg the color of animals.
This shows how different layers of an expert can be selected to obtain more
control over the synthesis process.

In Fig.~\ref{tab:sobeltoimg}, we consider the task of translating edge images
to natural images. Here, $\x$ is obtained through the Sobel filter, and we
choose a ResNet pretrained for image classification on stylized ImageNet as a
domain expert for edge images, as it has shown sensitivity to shapes
\cite{geirhos2018imagenet}.
This combination of $\oneenc$ and $\twodec$ through $\condinn$ enables
edge-to-image translation.
Fig.~\ref{tab:inpainting} shows an image inpainting task, where $\x$ is a
masked image. In this case, large portions of the shape are missing from the
image but the unmasked regions contain texture patches. This makes a ResNet
pretrained for image classification on ImageNet a suitable domain expert due to
its texture bias. The samples demonstrate that textures are indeed faithfully
preserved. 

Furthermore, we can employ the same approach for generative superresolution. 
Fig.~\ref{fig:superres} shows the resulting transfer when using our method for combining two
autoencoders, which are trained on different scales. More precisely, $\modelone$ is an autoencoder trained
on images of size $32 \times 32$, while $\modeltwo$ is an autoencoder of $256 \times 256$ images.
The samples show that the model captures the ambiguities \wrt this translation and thereby enables efficient superresolution.
\vspace{-0.15cm}
\superres
\paragraph{Model Diagnosis:}
\landscapeslayers
Besides being applicable for content creation, our approach to recovering the
invariances of $f$ can also serve as a diagnostic tool for model
interpretation. 
By comparing the generated samples $\y = \twodec(\reptwo)$ (see
Eq.~\eqref{eq:sample}) conditioned on representations $\repone = \oneenc(\x)$
extracted from \emph{different layers} of $\modelone$, we see how the invariances increase
with increasing layer depth and can thereby visualize what the model has
learned to ignore. Using the same segmentation model $f$ and autoencoder $g$ as
described above, we visualize the invariances and model representations for
different layers on the \emph{Animals} and a web-scraped \emph{Landscapes}
dataset. For the latter, Fig.~\ref{fig:landscapeslayers} demonstrates how the
model $f$ has learned to discard information in later layers; \eg the
variance in the synthesized outputs increases (such as different lightings or
colors for the same scene). The corresponding experiment on the \emph{Animals}
dataset is presented in Sec.~\ref{suppsec:ablation}.
There, we also study the importance of faithfully modeling ambiguities of the translation
by replacing the cINN with an MLP,
which in contrast to the cINN fails to translate deep representations.

Note that all results in Fig.~\ref{fig:animaltable} and
Fig.~\ref{fig:landscapeslayers},~\ref{fig:animalswap} were obtained by
combining a single, generic autoencoder $\modeltwo$, which has no capabilities
to process inputs of $\domainone$ on its own, and different domain experts
$\modelone$, which possess no generative capabilities at all. These results
demonstrate the feasibility of solving a wide-range of image-to-image tasks
through the fusion of pre-existing, task-agnostic experts on image domains
$\domainone, \domaintwo$.  Moreover, choosing different layers of the expert
$\modelone$ provides additional, fine-grained control over the generation
process.

\subsection{Evaluating image modification capabilities of our generic approach}
\label{subsec:imgmod}
\celebtab
\paragraph{Attribute Modification:}
To compare our generic approach against task-specific approaches, we compare its ability for
attribute modification on face images to those of \cite{choi2018stargan}. We
train the same autoencoder $\modeltwo$ as in the previous section on CelebA \cite{liu2015faceattributes}, and directly use
attribute vectors as $\repone$.  For an input image $\y$ with
attributes $\repone$, we synthesize versions with modified
attributes $\repone^*$. In each column of
Fig.~\ref{fig:celebtab}, we flip the binary entry of the corresponding
attribute to obtain $\repone^*$. To obtain the modified image, we
first compute $\reptwo = \twoenc(\y)$ and use its corresponding
attribute vector $\repone$ to obtain its attribute invariant representation $\modelinv =
\condinn^{-1}(\reptwo \vert \repone)$. We then mix it
again with the modified attribute vector to obtain $\reptwo^* =
\condinn(\modelinv \vert \repone^*)$, which can be readily
decoded to the modified image $\y^* = \twodec(\reptwo^*)$.

Qualitative results in Fig.~\ref{fig:celebtab} demonstrate successful
modification of attributes. In comparison to \cite{choi2018stargan}, our approach
produces more coherent changes, \eg changing gender causes changes in hair
length and changes in the beard attribute have no effect on female faces.
This demonstrates the advantage of fusing attribute information on a
low-dimensional representation of a generic autoencoder.
Overall, our approach produces images of higher quality, as
demonstrated by the FID scores \cite{heusel2017gans} in Fig.~\ref{fig:celebtab}.
Note that FID-scores are calculated \wrt the complete dataset, explaining the
high FID scores for attribute \emph{glasses}, where images consistently
possess a large black area.

\vspace{-0.4cm}
\paragraph{Exemplar-Guided Translation:}
\humananimalswap
Another common image modification task is exemplar-guided
image-to-image translation \cite{park2019semantic}, where the semantic content and
spatial location is determined via a label map and the style via an exemplar
image. To approach this task, we utilize the same segmentation model and
autoencoder as in Sec.~\ref{subsec:reuse}. As before, we use the last layer of
the segmentation model to represent semantic content and location. For a given
exemplar $\y$, we can then extract its residual $\modelinv =
\condinn^{-1}(\twoenc(\y) \vert \oneenc(\y))$ and the 
segmentation representation $\repone = \oneenc(\x)$ of another image
$\x$. Due to the independence of $\modelinv$ and $\repone$, we can now transfer
$\repone$ under the guidance of $\modelinv$ to a recombined $\reptwo^* =
\condinn(\modelinv \vert \repone)$ which is readily decoded to an image $\y^* =
\twodec(\reptwo^*)$ as shown in Fig.~\ref{fig:animalswap}.
\vspace{-0.25cm}
\paragraph{Unsupervised Disentangling of Shape and Appearance:}
Our approach also allows for an unsupervised variant of the previous task,
as shown in
Fig.~\ref{fig:unsupswap}. Here, we use a random spatial deformation $d$
(see Sec.~\ref{supp:unsupervised}) to define $\oneenc=\twoenc\circ d$. Thus,
$\repone$ and $\reptwo$ share the same appearance but differ in pose, which is
distilled into $\modelinv$. For exemplar
guided synthesis, the roles of $\x$ and $\y$ are now swapped.
\vspace{-0.25cm}
\paragraph{Unpaired Image Translation:}
Fig.~\ref{fig:secondpage} demonstrates results of our approach
applied to unpaired image-to-image translation. Here, we use the same setup as
for attribute modification, but train various cINN models for unpaired transfer between the following datasets: CelebA and AnimalFaces-HQ \cite{choi2019stargan}, FFHQ and CelebA-HQ \cite{karras2018progressive}, Anime \cite{danbooru2019Portraits} to CelebA-HQ/FFHQ and Oil Portraits to CelebA-HQ/FFHQ. Details regarding the training procedure can be found in the Sec.~\ref{supp:unpaired}.
\section{Conclusion}
\vspace{-0.25cm}
This paper has addressed the problem of generic domain transfer between
arbitrary fixed off-the-shelf domain models. We have proposed a conditionally
invertible domain translation network that faithfully transfers between
existing domain representations without altering them. Consequently, there is
no need for costly or potentially even infeasible retraining or finetuning of
existing domain representations. The approach is \emph{(i) flexible:} Our cINN
for translation as well as its optimization procedure are independent from the
individual domains and, thus, provide plug-and-play capabilities by allowing to
plug in arbitrary existing domain representations; \emph{(ii) powerful:}
Enabling the use of pretrained expert domain representations outsources the
domain specific learning task to these models. Our model can thus focus on the
translation alone which leads to improvements over previous approaches;
\emph{(iii) convenient and affordable}: Users can now utilize powerful,
pretrained models such as BERT and BigGAN for new tasks they were not designed
for, with just a single GPU instead of the vast multi-GPU resources required
for training such domain models. Future applications include transfer between
diverse domains such as speech, music or brain signals.

\creativityfigure
\FloatBarrier
\clearpage

\section*{Broader Impact}

\begin{description}
\item[Environmental and Economic Aspects:]\hfill\\ Single training runs of large scale
    models have a large environmental footprint due to the massive computational
    requirements. It is therefore unreasonable to repeat this effort for every
    new application. Instead we allow to reuse powerful models, leading to significant reductions in computational demands.
  \item[Boosting serendipity:]\hfill\\ New (scientific) knowledge arises where seemingly unrelated entities are brought together to study their relationship (cf. the 'double projection' proposed by Heinrich Wölfflin a century ago for art history and other image disciplines to easily contextualize diverse imagery of different cultures). Allowing to efficiently connect expert models for diverse data and problems, thus promises to provide the basis for new directions of future research.
  \item[Interdisciplinary research:]\hfill\\ One of the main stumbling blocks for interdisciplinary research (e.g. vision and language) is to bring together expert models for widely different domains. State-of-the-art models are typically developed in and for the individual disciplines. Being able to efficiently combine these disciplinary expert models to solve interdisciplinary problems promises to be an enabling factor for more effective cross-disciplinary research.
  \item[Social equality:]\hfill\\ Training state-of-the art models is typically so costly that only wealthy institutions can afford their transfer and application to different input domains or other subsequent research that would require to retrain them. Computationally efficient transfer of existing models with no need for costly retraining therefore increases the opportunities for economically weaker institutions and countries to have research programs in this field.
  \item[Increasing applicability and impact of research output:]\hfill\\ Large scale research is often funded by public resources and thus there
    is a responsibility to make results accessible to the public. While
    pretrained models are often shared publicly, the scope in which they can be
    applied is significantly widened by our approach.
  \item[Content creation and manipulation:]\hfill\\ Domain transfer applied to controlled image synthesis and modification has wide applicability in the creative industry and beyond. However, it can also be misused for forgery and manipulation.
\end{description}

\begin{ack}
  This work has been supported in part by the German Research Foundation (DFG)
  projects 371923335, 421703927, and EXC 2181/1 - 390900948, the German federal
  ministry BMWi within the project ``KI Absicherung'' and a hardware donation
  from NVIDIA Corporation.
\end{ack}

\FloatBarrier
\clearpage
{\small
\bibliographystyle{ieee_fullname}
\bibliography{ms}
}

\FloatBarrier
\clearpage
\appendix
\setcounter{page}{1}
\renewcommand{\thefigure}{S\arabic{figure}}
\setcounter{figure}{0}
\renewcommand{\thetable}{S\arabic{table}}
\setcounter{table}{0}

\begin{center}
  \textbf{
    \Large Network-to-Network Translation with \\\vspace{0.3em}Conditional Invertible Neural
  Networks} \\
  \Large 
-- \\
 \textbf{\large Supplementary Material} \\
\hspace{1cm}
\end{center}

\section*{Overview}
This appendix provides supplementary material for our work \emph{Network-to-Network Translation with cINNs}. Firstly, in Sec.~\ref{suppsec:energy}, we discuss and compare computational requirements and resulting costs of our model to other approaches, \eg BigGAN. Next, in Sec.~\ref{suppsec:objective}, we provide a derivation of the training objective (see Eq.\eqref{eq:loss}). Its interpretation as an information bottleneck allows for unsupervised disentangling of shape and appearance, which is demonstrated in Sec.~\ref{supp:unsupervised} and Fig.~\ref{supp:unsupswap}.
Sec.~\ref{supp:unpaired} then provides additional examples (Fig.~\ref{fig:suppswap}) and technical details for the unpaired image translation as presented in Fig.~\ref{fig:secondpage} (Sec.~\ref{subsec:imgmod}).
Subsequently, we continue with an ablation study in Sec.~\ref{suppsec:ablation}, where we analyze the performance of our approach by replacing the conditional invertible neural network with a multilayer perceptron. 
Next, Sec.~\ref{suppsec:aearch} presents the architecture and training procedure of our reusable autoencoder model $\modeltwo$, which we used to obtain the results shown in Fig.~\ref{fig:animaltable},~\ref{fig:landscapeslayers},~\ref{fig:animalswap},~\ref{fig:unsupswap},~\ref{supp:unsupswap},~\ref{fig:suppanimallayers},~\ref{fig:suppanimalswap}~and~\ref{fig:supplandscapesamples}.
Finally, Sec.~\ref{suppsec:implementationdetails} provides details on (i) the architecture of the cINN and (ii) the translation from BERT to BigGAN, \cf Sec.~\ref{subsec:transtobig}.

\section{Computational Cost and Energy Consumption}
\label{suppsec:energy}
In Tab.~\ref{tab:energy} we compare computational costs of our cINN to those of
BERT\footnote{BERT requirements: \url{https://ngc.nvidia.com/catalog/resources/nvidia:bert\_for\_tensorflow/performance}},
BigGAN\footnote{BigGAN
requirements: \url{https://github.com/ajbrock/BigGAN-PyTorch}} and
FUNIT.\footnote{FUNIT requirements: \url{https://github.com/NVlabs/FUNIT/}} The
Table shows that, once strong domain experts are available, they can be
repurposed by our approach in a time-, energy- and cost-effective way.  With
training costs of our cINN being two orders of magnitude smaller than the
training costs of the domain experts, the latter are amortized over all the new
tasks that can be solved by recombining experts with our approach.

\energyusage
\footnotetext[4]{Titan X Specs: \url{https://www.nvidia.com/en-us/geforce/products/10series/titan-x-pascal/}}
\footnotetext[5]{DGX-1 Specs: \url{https://docs.nvidia.com/dgx/dgx1-user-guide/introduction-to-dgx1.html}}
\footnotetext[6]{Electricity Prices: \url{https://ec.europa.eu/eurostat/statistics-explained/index.php?title=Electricity\_price\_statistics}}
\footnotetext[7]{CO2 Emission Intensity: \url{https://www.eea.europa.eu/data-and-maps/daviz/co2-emission-intensity-5}}

\section{Training Objective}
\label{suppsec:objective}
\paragraph{Derivation of Eq.~\eqref{eq:loss}:}
For a given $\repone$, we use a change of variables, $\modelinv =
\condinn^{-1}(\reptwo \vert \repone)$, to express the KL divergence with an
integral over $\reptwo$:
\begin{align}
  \KL(p(\modelinv \vert \repone) \vert q(\modelinv)) &= \int_{\modelinv}
  p(\modelinv \vert \repone) \log \frac{p(\modelinv \vert
  \repone)}{q(\modelinv)} \\
  &= \int_{\reptwo} p(\condinn^{-1}(\reptwo \vert \repone) \vert \repone)
  \vert \det J_{\condinn^{-1}}(\reptwo \vert \repone) \vert
  \log \frac{p(\condinn^{-1}(\reptwo \vert \repone) \vert
  \repone)}{q(\condinn^{-1}(\reptwo \vert \repone))} \label{eq:klovertwo}
\end{align}
By definition of $\modelinv$ in Eq.~\eqref{eq:induced}, the invertibility of
$\condinn$ allows us to express the conditional
probability density function $p(\modelinv \vert \repone)$ of $\modelinv$ in
terms of the conditional probability density function $p(\reptwo \vert
\repone)$ of $\reptwo$:
\begin{equation}
  p(\modelinv \vert \repone) = p(\condinn(\modelinv \vert \repone) \vert
  \repone) \vert \det J_{\condinn}(\modelinv \vert \repone) \vert
\end{equation}
For $\modelinv = \condinn^{-1}(\reptwo \vert \repone)$, the inverse function theorem then implies
\begin{equation}
  p(\condinn^{-1}(\reptwo \vert \repone) \vert \repone)
  = p(\reptwo \vert \repone) 
  \vert \det J_{\condinn^{-1}}(\reptwo \vert \repone) \vert^{-1}
  \label{eq:changeofvardensity}
\end{equation}
Using Eq.~\eqref{eq:changeofvardensity} in Eq.~\eqref{eq:klovertwo} gives
\begin{align}
  \KL(p(\modelinv \vert \repone) \vert q(\modelinv))
  &= \int_{\reptwo} p(\reptwo \vert \repone) \log \frac{p(\reptwo \vert
  \repone)}{q(\condinn^{-1}(\reptwo \vert \repone))\vert \det
  J_{\condinn^{-1}}(\reptwo \vert \repone) \vert} \\
  &= 
  \expect_{\reptwo} \left\{
    - \log q(\condinn^{-1}(\reptwo \vert \repone))
    - \log \vert \det J_{\condinn^{-1}}(\reptwo \vert \repone) \vert
    + \log p(\reptwo \vert \repone)
    \right\}
\end{align}
Taking the expectation over $\repone$ results in Eq.~\eqref{eq:loss}.
\newpage
\section{Unsupervised Disentangling of Shape and Appearance}
\label{supp:unsupervised}
\suppunsupswap
Unsupervised disentangling of shape and appearance aims to recombine shape, \ie
the underlying spatial structure, from one image with the appearance, \ie the
style, of another image. In contrast to Unpaired Image-to-Image translation of
Sec.~\ref{supp:unpaired}, training data does not come partitioned into a
discrete set of different image domains, and in contrast to Exemplar-Guided
Translation of Sec.~\ref{subsec:imgmod}, the task assumes that no shape expert,
\eg a segmentation model, is available. To handle this setting, we use the
encoder $\twoenc$ of the autoencoder trained on \emph{Animals} also for
$\oneenc$, but always apply a spatial deformation to its inputs. Thus, the
pairs $(\x, \y)$ are given by $(d(\y), \y)$, where $\y$ are the original images
from the \emph{Animals} dataset and $d$ is a random combination of horizontal
flipping, a thin-plate-spline transformation and cropping. The
translation task then consists of the translation of deformed encodings
$\repone = \oneenc(d(\y))\coloneqq \twoenc(d(\y))$ to the original encoding $\reptwo = \twoenc(\y)$.
After training, we apply our translation network to original images without the
transformation $d$. For two images $\x$ and $\y$, we obtain a shape
representation of $\y$ from its residual $\modelinv = \condinn^{-1}(\twoenc(\y)
\vert \twoenc(\y))$ and recombine it with the appearance of $\x$ to obtain $y^*
= \twodec(\condinn(\modelinv \vert \twoenc(\x)))$.
\supphumananimalswap
\suppunpairedtranslation
The results in Fig.~\ref{supp:unsupswap} demonstrate that our translation
network succesfully learns a shape representation $\modelinv$, which is
independent of the appearance and can thus be recombined with arbitrary
appearances, see Eq.~\eqref{eq:infobound}. Being able to operate completely unsupervised demonstrates the generality of our approach.
\newpage
\section{Unpaired Image Translation}
\label{supp:unpaired}
Unpaired Image-to-Image translation considers the case where only unpaired
training data is available. Following \cite{Zhu_2017}, let $Y^0=\{\y^0_i\}_{i=1}^N$
be a source set, and $Y^1=\{\y^1_j\}_{j=1}^M$ a target set. The goal is then to
learn a translation from source set to target set, with no information provided
as to which $\y^0_i$ matches which $\y^1_j$. We formulate this task as a
translation from a set indicator $x\coloneqq\repone \in \{0, 1\}$ to an output $y$, such that
for $\repone=0$, $y$ belongs to the source set $Y^0$, and for $\repone=1$, $y$ belongs to the
target set. Thus, in the case of unpaired image translation, the domains are
given by $\domainone = \{0, 1\}$ and $\domaintwo = Y^0 \cup Y^1$, with training pairs
$\{(0, y^0_i) \vert i=0,\dots,N\} \cup \{(1, y^1_j) \vert j=0,\dots,M\}$.
Because the residual $\modelinv$ is independent of $\repone$, it captures
precisely the commonalities between source and target set and therefore
establishes meaningful correspondences between them. To translate $\y^0_i$, we
first obtain its residual $\modelinv = \condinn^{-1}(\twoenc(\y^0_i) \vert 0)$,
and then decode it as an element of the target set $y^* = \twodec(\condinn(\modelinv
\vert 1))$. Note that the autoencoder $\modeltwo = \twodec \compose \twoenc$ is always trained on the 
combined domain $\domaintwo = Y^0 \cup Y^1$, \ie a combination of the two datasets of interest.

Additional examples for unpaired translation as in Fig.~\ref{fig:secondpage} can be found in Fig.~\ref{fig:suppswap}.
One can see that the viewpoint of a face is preserved upon translation, which
demonstrates that $\modelinv$ learns semantic correspondences between the pose
of faces from \emph{different} dataset modalities, without any paired data between them. The same holds for the samples in Fig.~\ref{fig:suppunpaired}, where the same $\modelinv$ is projected onto different domains $\domainone$.
\newpage

\section{Ablation Study: Replacing our cINN with an MLP}
\label{suppsec:ablation}
\suppanimallayers
To illustrate the importance of the cINN to model ambiguities of the
translation, we demonstrate the effect of replacing our cINN with a (deterministic) multilayer perceptron (MLP) in Fig.~\ref{fig:suppanimallayers}. The MLP consists of two parts: (i) an embedding part as in Tab.~\ref{tab:subembeddertab} and (ii) the architecture of a fully-connected network which was recently used for neural scene rendering \cite{mildenhall2020nerf}. We perform the same model diagnosis experiment as in Fig.~\ref{fig:landscapeslayers}, but applied to the
\emph{Animals} dataset. For early layers, the translation contains almost no
ambiguity and can be handled successfully by both the cINN and the MLP. For
a deeper layer of the used segmentation model $\modelone$, the translation
has moderate ambiguities, as $\modelone$'s invariances \wrt to the input increase. This is not accurately reflected by the multilayer perceptron, because it does not model the space of these invariances. Finally, for the last layer of $\modelone$, the MLP predicts the mean over all possible translation outputs which, due to its large ambiguities, does not
result in a meaningful translation anymore, whereas our cINN still samples
coherent translation outputs. FID scores in Tab.~\ref{fig:suppanimallayers} further validate this behavior for the whole test set.
\newpage
\section{Architecture and Training of the Autoencoder $\modeltwo$}
\label{suppsec:aearch}
\newcommand{\discr}{\mathcal{D}}
\suppanimalswap
\supplandscapesamples
All experiments in Sec.~\ref{subsec:reuse}, the exemplar-guided image-to-image
translation in Sec.~\ref{subsec:imgmod} as well as the additional results in
Fig.~\ref{fig:suppanimalswap} and Fig.~\ref{fig:supplandscapesamples} were conducted using the \emph{same} (\ie same weights and architecture) autoencoder $\modeltwo$, thereby demonstrating how a single model can be re-used for multiple purposes within our framework. \\
For the autoencoder, we use a ResNet-101 \cite{he2016deep} architecture as encoder $\twoenc$, and the BigGAN architecture as the decoder $\twodec$, see Tab.~\ref{tab:aearchitecture}. As we do not use class information, we feed the latent code $\reptwo$ of the encoder into a a fully-connected layer and use its softmax-activated output as a replacement for the one-hot class vector used in BigGAN. The encoder predicts mean $\twoenc(\y)_\mu$ and diagonal covariance $\twoenc(\y)_{\sigma^2}$ of a Gaussian distribution and we use the
reparameterization trick to obtain samples $\reptwo = \twoenc(\y)_\mu + \diag(\twoenc(\y)_{\sigma^2}) \epsilon$ of the latent code, where $\epsilon \sim \normaldistr(0, \id)$. For the reconstruction loss $\Vert \cdot \Vert$, we use a perceptual loss based on features of a pretrained VGG-16 network \cite{simonyan2014very}, and, following \cite{dai2019diagnosing}, include a learnable, scalar output variance $\gamma$. Additionally, we use the PatchGAN discriminator $\discr$ from \cite{isola2017image} for improved image quality.
Hence, given the autoencoder loss
\begin{align}
  \loss_{VAE}(\twoenc, \twodec, \gamma)
  = \mathbb{E}_{\begin{subarray}{l} \y \sim p(\y)\\ \epsilon \sim \normaldistr(0, \id)\end{subarray}}
  &\bigg[ \frac{1}{\gamma}
  \Vert \y - \twodec(\twoenc_{\mu}(\y) +
  \sqrt{\diag(\twoenc_{\sigma^2}(\y))}\;\epsilon) \Vert + \log \gamma \nonumber  \\
&+\KL\big(\normaldistr\left(\reptwo \vert \twoenc_{\mu}(\y), \diag(\twoenc_{\sigma^2}(\y)\big) \Vert \normaldistr(0, \id) \right)  \bigg],
\label{loss:vae}
\end{align}
and the GAN-loss
\begin{align}
\loss_{GAN}(\modeltwo, \discr) = \mathbb{E}_{\begin{subarray}{l} \y \sim p(\y)\\ \epsilon \sim \normaldistr(0, \id)\end{subarray}} \left[\log \discr(y) + \log \left( 1 - \twodec\Big(\twoenc_{\mu}(\y) +
  \sqrt{\diag(\twoenc_{\sigma^2}(\y))}\;\epsilon\Big) \right) \right],
\end{align}
the total objective for the training of $\modeltwo = \{\twoenc, \twodec, \gamma\}$ reads:
\begin{align}
\{\twoenc^*, \twodec^*, \gamma^*\} = \arg \min_{\twoenc, \twodec, \gamma} \max_{\discr} \left[ \loss_{VAE}(\twoenc, \twodec, \gamma) + \lambda \loss_{GAN}(\{\twoenc, \twodec, \gamma\}, \discr) \right].
\end{align}
This is similar to the improved image metric suggested in \cite{dosovitskiy2016generating}, but in contrast to their work, we use an \emph{adaptive} weight $\lambda$, computed by the ratio of the gradients of the decoder $\twodec$ \wrt its last layer $\twodec_{L}$:
\begin{equation}
\lambda = \frac{\Vert \nabla_{\twodec_L}(\loss_{rec}) \Vert}{\Vert \nabla_{\twodec_L}(\loss_{GAN}) \Vert + \delta}
\end{equation}
where the reconstruction loss $\loss_{rec}$ is given as (\cf Eq.~\eqref{loss:vae}): 
\begin{equation}
\loss_{rec} = \mathbb{E}_{\begin{subarray}{l} \y \sim p(\y)\\ \epsilon \sim \normaldistr(0, \id)\end{subarray}}
  \bigg[ \frac{1}{\gamma}
  \Vert \y - \twodec(\twoenc_{\mu}(\y) +
  \sqrt{\diag(\twoenc_{\sigma^2}(\y))}\;\epsilon) \Vert + \log \gamma \bigg], 
\end{equation}  
and a small $\delta$ is added for numerical stability.
\aearchitecture
\newpage
\section{Implementation Details}
\label{suppsec:implementationdetails}
\subsection{Architecture of the conditional INN}
\label{subsec:cinnarch}
\flowblock
\minimodelstab
In our implementation, the conditional invertible neural network (cINN) consists of a sequence of INN-blocks as shown in Fig.~\ref{fig:flowblock}, and a non-invertible embedding module $H$ which provides the conditioning information. Each INN-block is build from (i) an alternating affine coupling layer \cite{dinh2016density}, (ii) an activation normalization (\emph{actnorm}) layer \cite{kingma2018glow}, and (iii) a fixed permutation layer, which effectively mixes the components of the network's input. Given an input $z \in \RR^D$ and additional conditioning information $y$, we pre-process the latter with a neural network $H$ as
\begin{equation}
h=H(y),
\label{eq:condembed}
\end{equation}
and a single \emph{conditional} affine coupling layer splits $z$ into two parts $z_{1:d}$, $z_{d+1:D}$ and computes
\begin{align}
z_{1:d}, z_{d+1:D} &= \text{\texttt{split}}(z) \\
z' &= \text{\texttt{concat}}\Big(z_{1:d}, s_{\theta}([z_{1:d}; h]) \odot z_{d+1:D} + t_{\theta}([z_{1:d};h])\Big)
\label{eq:singleaffine}
\end{align}
where $s_{\theta}$ and $t_{\theta}$ are implemented as simple feedforward neural networks, which process the concatenated input $[z_{i:j}, h]$ (see Tab.~\ref{tab:subbasicfully}). The \emph{alternating} coupling layer ensures mixing for all components of $z$:
\begin{align}
z'' &= \text{\texttt{concat}}\Big(s_{\theta}([z'_{d+1:D};h]) \odot z_{1:d} + t_{\theta}([z'_{d+1:D};h]), z'_{d+1:D}\Big).
\label{eq:altaffine}
\end{align}
We implement the embedding module $H$ as a simple (convolutional) neural network, see Tab.~\ref{tab:subembeddertab} for details regarding its architecture. Note that $H$ does \emph{not} need to be invertible as it soley processes conditioning information. Hence, given some $h=H(y)$, the network $\condinn$ is conditionally invertible. Usually, we train with a batch size of 10-25, which requires 4-12 GB VRAM and converges
in less than a day.

\subsection{Training Details for Bert-to-BigGAN Translation}
\label{subsec:berttobigdetails}
\berttobigflat
Training our approach to translate between a model's $\modelone$ representation $\repone = \oneenc(\x)$ and BigGAN's latent space $\reptwo$ requires to dequantize the discrete class labels $c$ that BigGAN is trained with. To do so, we consider the stacked vector 
\begin{equation}
\reptwo' =\left[\tilde{\modelrep}, Wc\right],
\label{eq:biggan}
\end{equation}
consisting of $\tilde{\modelrep} \sim \normaldistr(0, \id)$, $\tilde{\modelrep}
\in \RR^{140}$, sampled from a multivariate normal distribution and $c \in \{0,
1\}^K$, a one-hot vector specifying an ImageNet class ($K = 1000$ classes in
total). The matrix $W$, a part of the generator $\twodec$, maps the one-hot
vector $c$ to $h \in \RR^{128}$, \ie $h = Wc$, such that $\reptwo'$ in
Eq.~\eqref{eq:biggan} corresponds to a synthesized image, given a pretrained
generator of BigGAN. However, as $c$ contains discrete labels, we have to avoid
collapse of $\condinn$ onto a single dimension of $h$ during training. To this
end, we pass the vector $h$ through a small, fully connected variational
autoencoder (described in Tab.~\ref{tab:minivae}) and replace $h$ by its
stochastic reconstruction $\hat{h}$, which effectively performs dequantization,
such that:
\begin{equation}
\reptwo =\left[\tilde{\modelrep}, \hat{h}\right], \quad \text{with} \; \reptwo \in \RR^{268}.
\label{eq:bigganmod}
\end{equation}
Training of $\condinn$ is then conducted by sampling $\reptwo$ as in Eq.~\eqref{eq:bigganmod} and minimizing the objective described in Eq.~\eqref{eq:loss}, \ie finding a mapping $\condinn$ that conditionally maps $\repone$ and the corresponding ambiguities $\modelinv \sim \normaldistr(0,\id)$ to $\modeltwo$'s representations $\reptwo$. Additional results obtained when using this approach to conditionally translate BERT's representation $\repone$ into the latent space of BigGAN (see also Sec.~\ref{subsec:transtobig}) can be found in Fig.~\ref{fig:berttobigsupp}.
\suppminivae

\end{document}